\documentclass[lettersize,journal]{IEEEtran}

\usepackage{amsmath,amsfonts,amssymb}
\usepackage{amsthm}
\usepackage{algorithm}
\usepackage{algorithmic}
\usepackage{array}
\usepackage[caption=false,font=normalsize,labelfont=sf,textfont=sf]{subfig}
\usepackage{textcomp}
\usepackage{stfloats}
\usepackage{url}
\usepackage{verbatim}
\usepackage{graphicx}
\usepackage{balance}
\def\BibTeX{{\rm B\kern-.05em{\sc i\kern-.025em b}\kern-.08em
    T\kern-.1667em\lower.7ex\hbox{E}\kern-.125emX}}

\usepackage{microtype}
\usepackage[utf8]{inputenc} 
\usepackage[T1]{fontenc}    
\usepackage{nicefrac}       
\usepackage{multirow}
\usepackage{booktabs}       
\usepackage{xcolor}
\usepackage{color, colortbl}
\usepackage[normalem]{ulem}
\usepackage{arydshln}
\usepackage{enumitem}
\usepackage{wrapfig}
\usepackage{bbm}
\usepackage{bm}
\usepackage[calc]{adjustbox}
\usepackage{tabularx}
\usepackage{makecell}
\usepackage{gensymb}
\usepackage{marvosym}

\renewcommand{\algorithmiccomment}[1]{\bgroup\hfill$\triangleright$~#1\egroup}

\makeatletter
\def\hlinewd#1{%
	\noalign{\ifnum0=`}\fi\hrule \@height #1 %
	\futurelet\reserved@a\@xhline}
\makeatother

\definecolor{Red}{rgb}{1,0,0}
\definecolor{Blue}{rgb}{0,0,0.8}
\definecolor{Green}{rgb}{0,0.7,0.2}
\definecolor{airforceblue}{rgb}{0.36, 0.54, 0.66}
\definecolor{ao(english)}{rgb}{0.0, 0.5, 0.0}
\definecolor{azure(colorwheel)}{rgb}{0.0, 0.5, 1.0}
\definecolor{crimson}{rgb}{0.86, 0.08, 0.24}
\definecolor{darkcerulean}{rgb}{0.03, 0.27, 0.49}
\definecolor{cobalt}{rgb}{0.0, 0.28, 0.67}
\definecolor{rosegold}{rgb}{0.72, 0.43, 0.47}
\definecolor{orange-red}{rgb}{1.0, 0.27, 0.0}
\definecolor{mountainmeadow}{rgb}{0.19, 0.73, 0.56}
\definecolor{malachite}{rgb}{0.04, 0.85, 0.32}
\definecolor{darkblue}{rgb}{0.0, 0.0, 0.55}
\definecolor{customblue}{rgb}{0.2, 0.35, 0.8}
\definecolor{gg}{gray}{0.9}

\newcommand{\eat}[1]{{}}

\newtheorem{theorem}{Theorem}[section]
\newtheorem{corollary}{Corollary}[theorem]
\newtheorem{lemma}[theorem]{Lemma}
\newtheorem{proposition}[theorem]{Proposition}

\usepackage[numbers,sort&compress]{natbib}

\usepackage{hyperref}
\hypersetup{colorlinks=true}
\hypersetup{linktoc=all}
\hypersetup{citecolor=customblue}
\hypersetup{linkcolor=crimson}
\hypersetup{urlcolor=darkblue}
\usepackage[all]{hypcap}

\usepackage[nameinlink]{cleveref}
\creflabelformat{equation}{#2\textup{#1}#3}
\crefname{assumption}{assumption}{assumptions}

\begin{document}

\title{Soft-TransFormers for Continual Learning}

\author{Haeyong~Kang and~Chang~D.~Yoo%
\thanks{Manuscript received July 16, 2026; revised XX XX, 2026. \textit{(Corresponding author: Haeyong Kang.)}}%
\thanks{Haeyong Kang and Chang D. Yoo are with the School of Electrical Engineering, Korea Advanced Institute of Science and Technology (KAIST), Daejeon 34141, South Korea (e-mail: haeyong.kang@kaist.ac.kr).}}

\markboth{IEEE Transactions on Neural Networks and Learning Systems,~Vol.~XX, No.~X, July~2026}%
{Kang \MakeLowercase{\textit{et al.}}: Soft-TransFormers for Continual Learning}

\maketitle

\begin{abstract}
Inspired by the Well-initialized Lottery Ticket Hypothesis (WLTH), we introduce Soft-TransFormers (Soft-TF), a continual learning framework that adapts a frozen pre-trained Transformer through task-specific \emph{soft subnetworks}: real-valued multiplicative masks over the query, key, value, and output projections of selected self-attention layers. The masks are initialized at one, so optimization starts exactly at the pre-trained solution, and mask-space gradient descent is intrinsically biased toward modulating the backbone's dominant pathways; we prove that, under standard convex-Lipschitz assumptions, both the convergence rate and the parameter drift of mask-only fine-tuning are controlled by the distance from the pre-trained weights to a task-optimal configuration. This bounded drift yields two properties. Since the backbone and per-task masks are never overwritten, forgetting is structurally eliminated. And since every task subnetwork stays near the shared pre-trained solution, a wrong mask still evaluates a near-generalist function, so task-inference errors are largely harmless and class-incremental accuracy is decoupled from task-inference reliability. As a plug-in, Soft-TF couples with L2P, DualPrompt, HiDe-Prompt, and NoRGa, selecting masks by task-key matching, an entropy-gradient criterion, or a learned task-identity classifier. Across class-incremental learning benchmarks---Split-CIFAR100, Split-ImageNet-R, CUB-200, and 5-Datasets---Soft-TF consistently outperforms prompt-based, adapter-based, and LoRA-style baselines at comparable trainable-parameter budgets, while keeping inference cost identical to the unmodified backbone.
\end{abstract}


\section{Introduction}

Continual Learning (CL), also known as Lifelong Learning \cite{ThrunS1995, rusu2016progressive, zenke2017continual, hassabis2017neuroscience}, aims to endow models with the ability to acquire new knowledge over time without erasing previously learned information. However, modern neural networks remain highly susceptible to \emph{catastrophic forgetting}~\cite{McCloskey1989}: parameter updates for new tasks interfere with the parameters that encode earlier ones, resulting in severe performance degradation. Classical remedies developed for convolutional architectures fall into regularization-based, rehearsal-based, and architecture-based families; we defer a detailed discussion to \Cref{sec:related}.

\begin{figure*}[ht]
    \centering
    \small
    \makebox[\textwidth]{%
        \includegraphics[height=4.3cm, trim={1.cm 1.cm 0.0cm 1.cm},clip]{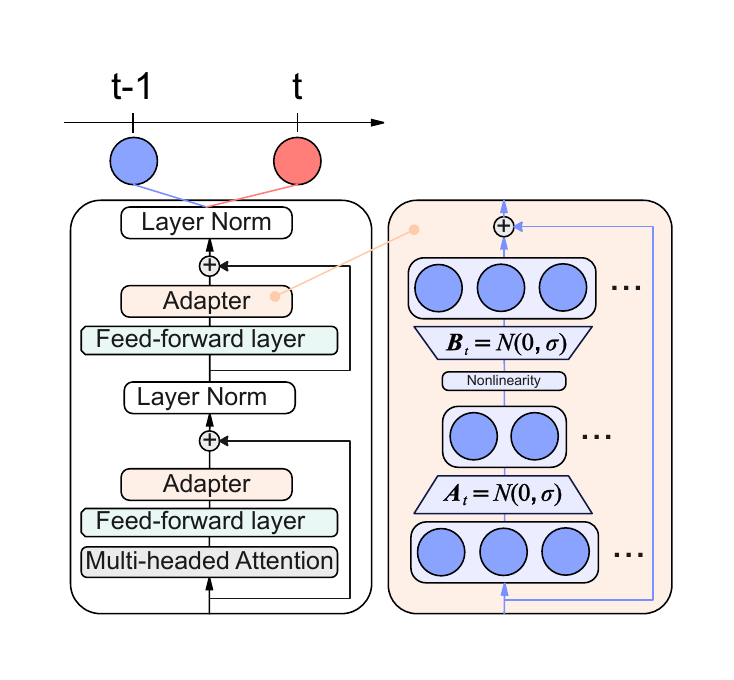}\hfill
        \includegraphics[height=4.3cm, trim={1.cm 1.cm 0.0cm 1.cm},clip]{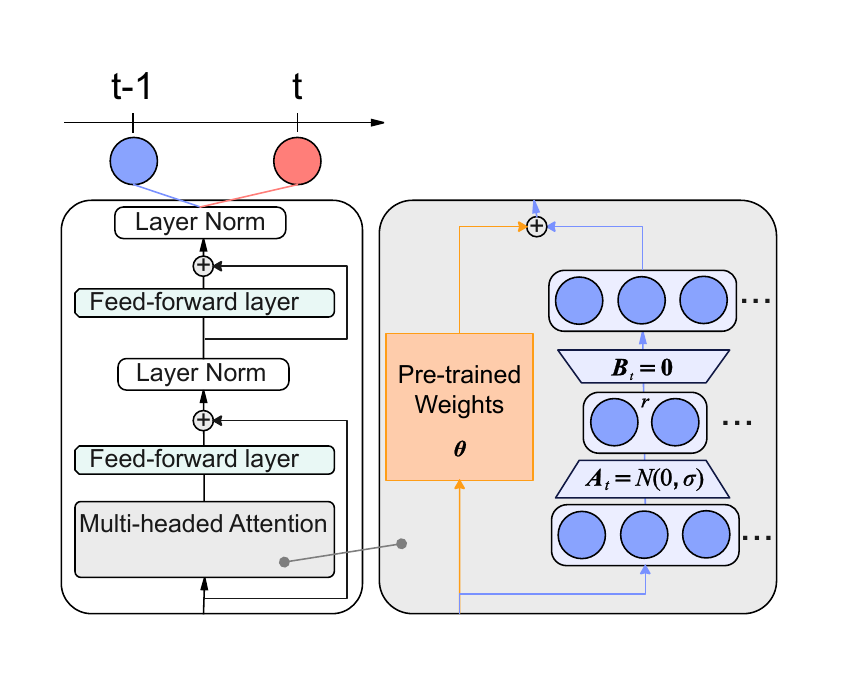}\hfill
        \includegraphics[height=4.3cm, trim={1.cm 1.cm 0.0cm 1.cm},clip]{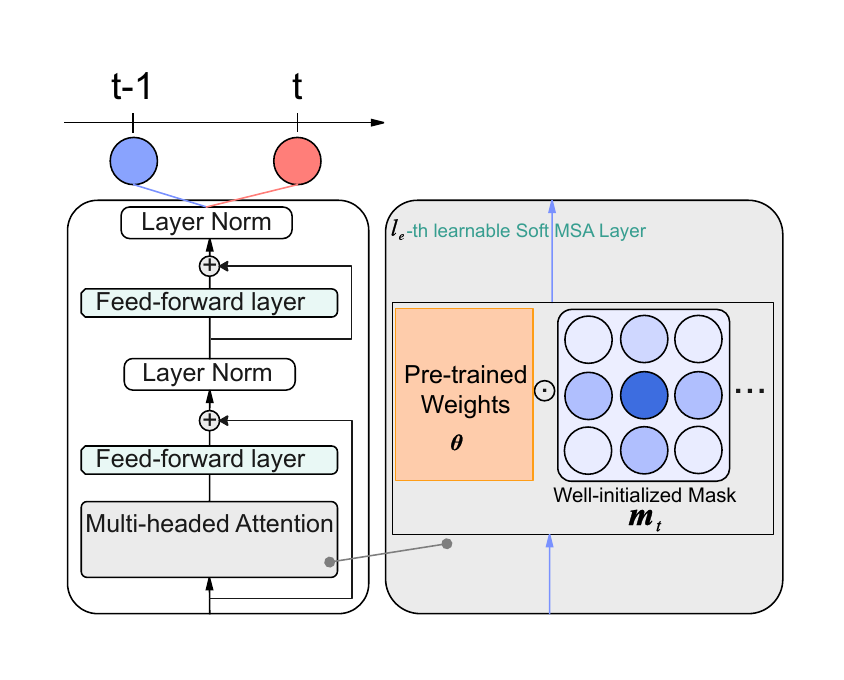}}\\[2pt]
    \makebox[\textwidth]{%
        \makebox[0.333\textwidth]{\small (a) Adapter}%
        \makebox[0.333\textwidth]{\small (b) LoRA}%
        \makebox[0.333\textwidth]{\small (c) our \textbf{Soft-TransFormer}}}
    \caption{\small \textbf{Comparisons of LLM-based fine-tuning methods (Adapters, LoRA) with Soft-TransFormer (Soft-TF)}: well-initialized Soft-TF $\bm{\theta} \odot \bm{m}_t$ is trained at $l_e$-th or only a few attention layers while LLM fine-tuning methods such as Adapter~\cite{houlsby2019parameter} and LoRA~\cite{hu2021lora} are trained at all attention layers. Note that all pre-trained parameters $\bm{\theta}$ are fixed in fine-tuning t-th task parameter $\bm{m}_t$ in Continual Learning (CL) scenarios.}
    \label{fig:llm_soft_tf}
    \vspace{-0.15in}
\end{figure*}

With the advent of large-scale pre-trained Transformers such as ViT \cite{dosovitskiy2020image}, \emph{prompt-based} continual learning has emerged as the dominant rehearsal-free paradigm: the backbone is frozen and only lightweight prompts are learned per task, beginning with L2P~\cite{wang2022learning} and DualPrompt~\cite{wang2022dualprompt} and followed by numerous refinements~\cite{Smith_2023_CVPR, wang2022s, wang2023HiDe, gao2024cprompt, le2025NoRGa}. Despite their efficiency, prompt-only methods share a structural limitation: prompts condition the frozen computation solely through the \emph{input} of each attention layer, so their representational flexibility is fundamentally lower than weight-level adaptation. As task sequences become longer and more heterogeneous, prompt pools grow, retrieval becomes brittle, and the restricted capacity of prompts fails to capture fine-grained task-specific variations.

To address these challenges, we revisit the parameter-efficient fine-tuning (PEFT) landscape—Adapters \cite{houlsby2019parameter}, LoRA \cite{hu2021lora}, and learned rescalings such as (IA)$^3$~\cite{liu2022tfew}—from a continual learning perspective, as illustrated in \Cref{fig:llm_soft_tf}. Two desiderata guide this re-examination of sequentially adapting a pre-trained foundation model: minimizing catastrophic forgetting and maintaining parameter efficiency throughout continual adaptation. These goals motivate the following hypothesis:

\noindent\textbf{Well-initialized Lottery Ticket Hypothesis (WLTH).}
\emph{A well-initialized pre-trained network already contains, near its frozen weights, soft real-valued subnetworks that specialize to new tasks without overwriting prior knowledge.}

\noindent
Inspired by WLTH, we propose \textbf{Soft-TransFormers (Soft-TF)}, a continual learning framework that discovers \emph{task-specific soft subnetworks} inside a frozen pre-trained Transformer, as shown in \Cref{fig:dual_soft}. For each task, Soft-TF learns real-valued multiplicative masks—initialized at one, so that optimization starts exactly at the pre-trained solution—that modulate the query, key, value, and output projections of selected self-attention layers. Crucially, Soft-TF is not a standalone architecture but a \emph{plug-in} mechanism: it composes with existing prompt-based learners such as L2P, DualPrompt, and NoRGa, consistently improving each of them.

\noindent
Our main contributions are as follows:
\begin{itemize}[leftmargin=*]
    \item \textbf{Method.} We introduce \emph{Soft-TransFormers (Soft-TF)}, which adapts a frozen pre-trained Transformer through task-specific real-valued masks over its attention projections, initialized at one so adaptation starts exactly at the pre-trained solution. It is a plug-in that composes with any prompt-based learner, unlike additive PEFT such as Adapters and LoRA or activation-rescaling PEFT such as (IA)$^3$.
    \item \textbf{Analysis.} We prove that mask-only training stays within a bounded drift of the pre-trained weights, and identify its two continual-learning consequences: forgetting is structurally eliminated, and class-incremental accuracy is decoupled from task-inference accuracy, since a mis-selected mask still evaluates a near-generalist function.
    \item \textbf{Results.} Soft-TF reaches state-of-the-art class-incremental performance as a plug-in to L2P, DualPrompt, HiDe-Prompt, and NoRGa across Split-CIFAR100, Split-ImageNet-R, CUB-200, and 5-Datasets; a controlled study verifies the decoupling, where a selector correct on only 58\% of samples still yields near-87\% accuracy.
\end{itemize}

\section{Related Work}\label{sec:related}

\noindent
\textbf{Continual Learning.} CL accumulates knowledge over a task sequence without forgetting earlier tasks~\cite{McCloskey1989,ThrunS1995}. Regularization-based methods constrain updates that would harm past tasks~\cite{Kirkpatrick2017,chaudhry2020continual,Jung2020,titsias2019functional,mirzadeh2020linear}; rehearsal-based methods replay stored or synthesized data~\cite{rebuffi2017icarl,riemer2018learning,chaudhry2018efficient,chaudhry2019continual,buzzega2020dark,Saha2021}; and architecture-based methods isolate disjoint parameter subsets per task~\cite{mallya2018piggyback,Serra2018,li2019learn,wortsman2020supermasks,kang2022forget}. Soft-TF is closest to the last family, especially to mask-based subnetworks such as Piggyback~\cite{mallya2018piggyback}, HAT~\cite{Serra2018}, supermasks~\cite{wortsman2020supermasks}, and WSN~\cite{kang2022forget}, later extended to video representations~\cite{kang2024fso,kang2024progressive}, and to SoftNet~\cite{kang2022soft}, which pairs a score-selected binary subnetwork with real-valued minor weights. Soft-TF differs in two ways: its masks are \emph{fully} soft, needing no discrete selection or scoring, and they modulate a \emph{frozen pre-trained} Transformer, which the Well-initialized Lottery Ticket Hypothesis (WLTH) posits already holds stable, task-adaptable subspaces.

\noindent
\textbf{Parameter-efficient Fine-tuning (PEFT).} PEFT adapts large pre-trained Transformers~\cite{dosovitskiy2020image,radford2021learning} with few trainable parameters: Adapters~\cite{houlsby2019parameter} insert small modules, LoRA~\cite{hu2021lora} adds low-rank \emph{additive} updates $\Delta\bm{W}=\bm{A}\bm{B}$, and (IA)$^3$~\cite{liu2022tfew} learns \emph{multiplicative rescaling vectors} on activations. Soft-TF shares the multiplicative principle of (IA)$^3$ but differs in three ways: it learns full element-wise masks over the weight matrices $\bm{\theta}^{Q,K,V,O}$ rather than rank-one activation vectors; it maintains a \emph{set} of per-task masks learned sequentially and selected at inference without task identifiers; and it is WLTH-derived, its mask-at-one initialization enabling the bounded-drift analysis of \Cref{sec:stability}. Unlike per-task LoRA, a Soft-TF mask of one reproduces the backbone exactly, and this well-initialized parameterization empirically outperforms additive low-rank updates in CIL, as shown in \Cref{sec:exp}.

\noindent
\textbf{Prompt-based Continual Learning.} Prompt-based CL adapts a frozen Transformer through learnable tokens. L2P~\cite{wang2022learning} retrieves prompts from a pool; DualPrompt~\cite{wang2022dualprompt} splits them into task-general and task-specific prefixes; and later work improves expressiveness and robustness, including PGP~\cite{qiao2024prompt}, CODA-Prompt~\cite{Smith_2023_CVPR}, S-Prompt~\cite{wang2022s}, HiDe-Prompt~\cite{wang2023HiDe}, CPrompt~\cite{gao2024cprompt}, and NoRGa~\cite{le2025NoRGa}. Since prompts condition only the input, their adaptation capacity is bounded, motivating Soft-TF's weight-level yet still frozen-backbone modulation.

\noindent
\textbf{Modular and Expert-Routing Continual Learning.} A parallel line equips frozen models with per-task modules \emph{routed} at inference: RanPAC~\cite{mcdonnell2023ranpac} accumulates class prototypes over random projections, mixture-of-experts adapters route to per-task experts~\cite{yu2024boosting}, synthetic data calibrates the router~\cite{byun2025expert}, dynamic rank-selective LoRA allocates capacity per task~\cite{lu2024adaptive}, and modality-gap methods address CLIP drift~\cite{huang2025mindgap}. Soft-TF stores and selects one mask set per task but adds \emph{no new modules}: adaptation stays inside the pre-trained weights, reproducing a full-capacity specialized backbone at unchanged inference cost. It is also uniquely tolerant of selector errors: because it starts \emph{at} the pre-trained solution and every task subnetwork stays near this shared backbone, a mis-routed input is still processed by a near-generalist function---unlike routed experts that drift far from any shared reference. This decouples class-incremental accuracy from selector accuracy, as we establish in \Cref{sec:stability} and verify in \Cref{sec:exp}.

\section{Preliminaries}\label{sec:prelim}
We first formalize the continual learning setting and review the prompt-based methods on which Soft-TF builds, fixing the notation used throughout the paper.

\subsection{Problem Statement}
Continual Learning (CL) trains a model on time-variant data represented as a sequence of tasks $\mathcal{D} = \{\mathcal{D}_1, \cdots, \mathcal{D}_\mathcal{T}\}$. The $t$-th task $\mathcal{D}_t =\{(\bm{x}_i^t, y_i^t)\}^{n_t}_{i=1}$ consists of $n_t$ pairs, where $\bm{x}_i^t \in \mathcal{X}_t$ is an input sample and $y_i^t \in \mathcal{Y}_t$ its label. When task $t$ arrives, the model is trained on $\mathcal{D}_t$ only; data from previous tasks is inaccessible. We focus on Class-Incremental Learning (CIL), where the task identity is \emph{not} given at inference time and predictions are made over all classes seen so far; this contrasts with Task-Incremental Learning (TIL), where the task identity is provided. Note that, as in standard CIL, task boundaries \emph{are} available during training; the challenge is test-time inference without them.

\subsection{Prompt-based Class-Incremental Learning}\label{sub_sec:prompt_cil}
Learning to Prompt (L2P)~\cite{wang2022learning} adapts a frozen pre-trained ViT $f_{\bm{\theta}}$ by prepending a small set of trainable tokens, called prompts, to the patch embeddings. L2P maintains a prompt pool of prompt--key pairs $\{(\bm{p}_t, \bm{k}_t)\}_{t=1}^\mathcal{T}$, where $\bm{p}_t \in \mathbb{R}^{L_p \times D}$ is a prompt of length $L_p$ and $\bm{k}_t \in \mathbb{R}^{D}$ is its learnable key. Given an input $\bm{x}$, a \emph{query function} $q(\bm{x})$—the [class] token feature of the frozen backbone applied to $\bm{x}$ without prompts—retrieves the prompt whose key is closest in cosine distance,
\begin{equation}\label{eq:key_match}
    \hat{t} = \operatorname*{argmin}_{t} \; \gamma\big(q(\bm{x}), \bm{k}_t\big),
    \qquad
    \gamma(\bm{u},\bm{v}) = 1 - \frac{\langle \bm{u}, \bm{v}\rangle}{\|\bm{u}\|\,\|\bm{v}\|},
\end{equation}
and the keys are trained by minimizing $\gamma(q(\bm{x}), \bm{k}_t)$ on task-$t$ data so that queries of task $t$ cluster around $\bm{k}_t$.

DualPrompt~\cite{wang2022dualprompt} refines L2P in two ways. First, it decomposes prompts into a single \emph{general} (G-) prompt $\bm{g}$, shared across all tasks and capturing task-invariant knowledge, and per-task \emph{expert} (E-) prompts $\{\bm{e}_t\}_{t=1}^{\mathcal{T}}$ capturing task-specific knowledge. Second, it replaces prompt-tuning with \emph{prefix-tuning}: rather than being prepended to the input sequence, prompts are split and attached to the key and value inputs of selected multi-head self-attention (MSA) layers. At test time, the E-prompt is selected by the same key-matching rule as \Cref{eq:key_match}. Since Soft-TF inherits this retrieval mechanism—using it to select not only the E-prompt but also the task-specific soft masks—we make the mechanism and its failure modes explicit in \Cref{sub_sec:task_inference}.

\begin{figure*}[!ht]
    \centering
    \setlength{\tabcolsep}{-1pt}{%
    \includegraphics[width=0.99\textwidth, trim={1.0cm 1.0cm 1.0cm 1.0cm},clip]{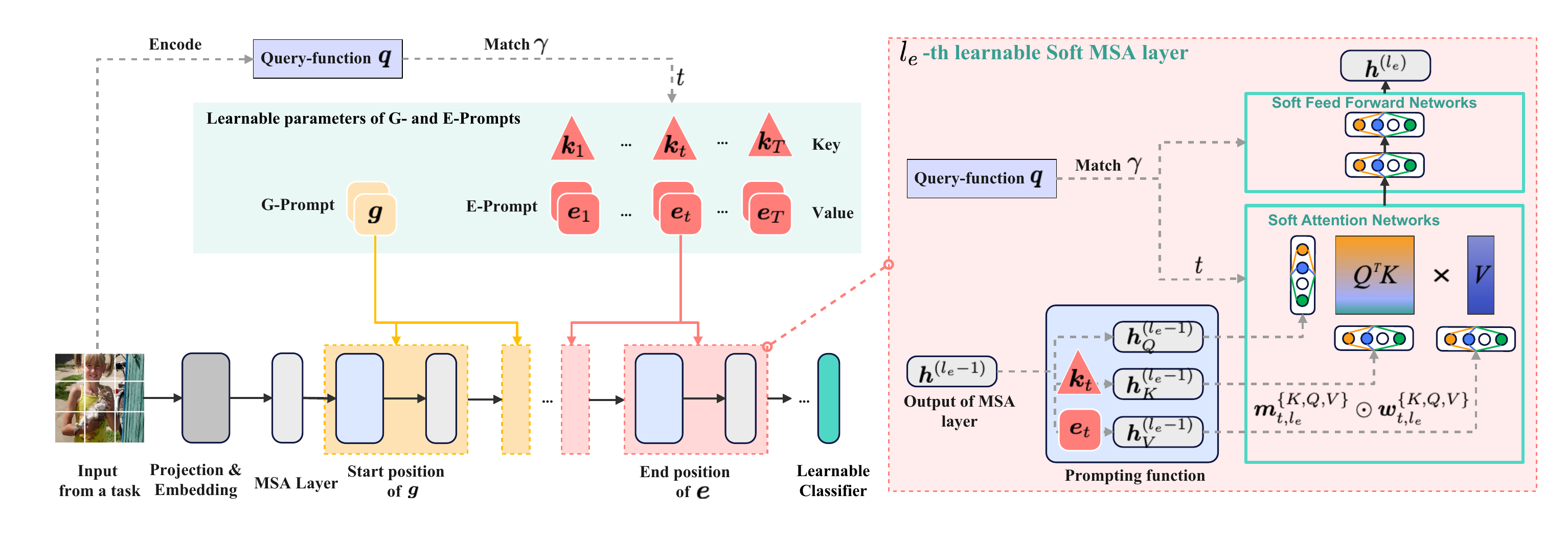}
    }
    \caption{\small \textbf{Soft-TransFormer (Soft-TF)}: At training time, the E-Prompt and the Soft-network are selected according to task identity, and the selected G-Prompt, E-Prompt, and the Soft-MSAs are trained together with a classifier. At test time, an input is transformed by a query function (Prompt ID) or task identifier to match the closest task key $\bm{k}_t$, E-prompt $\bm{e}_t$ and Soft-networks $\bm{m}_{t}^{\{Q,K,V,O\}}$.}
    

    \label{fig:dual_soft}
    \vspace{-0.15in}
\end{figure*}

\section{Soft-TransFormers (Soft-TF)}\label{sec:method}
This section presents Soft-TransFormers (Soft-TF), a continual learning framework that adapts a frozen pre-trained Transformer through continuous, learnable \emph{soft subnetworks} coupled with structured multi-layer prompting. \Cref{sub_sec:soft_msa} defines the soft multi-head self-attention layer; \Cref{sub_sec:coupling,sub_sec:objective,sub_sec:task_inference,sub_sec:cost} describe its coupling with dual prompts, the training objective, test-time task inference, and parameter cost; \Cref{sec:stability} closes with an optimization analysis of the induced parameter drift.

\subsection{Soft Multi-head Self-Attention}\label{sub_sec:soft_msa}
Let $\bm{\theta}$ denote the parameters of the pre-trained Transformer, which remain frozen throughout continual learning, and let $\bm{\theta}^Q, \bm{\theta}^K, \bm{\theta}^V, \bm{\theta}^O \in \mathbb{R}^{D \times D}$ be the query, key, value, and output projections of an MSA layer. For each task $t$, Soft-TF introduces real-valued masks of the same shapes,
\begin{equation}\label{eq:masks}
\bm{m}_t = \{\bm{m}_t^Q,\, \bm{m}_t^K,\, \bm{m}_t^V,\, \bm{m}_t^O\},
\qquad
\bm{m}_t^{(1)} = \mathbf{1},
\end{equation}
which modulate the projections element-wise through the Hadamard product $\odot$, yielding the \emph{effective} task-$t$ weights
\begin{equation}\label{eq:effective}
\bm{w}_t \;=\; \bm{\theta} \odot \bm{m}_t .
\end{equation}
Let $\bm{z} = [\,\bm{p};\, \bm{x}\,] \in \mathbb{R}^{D \times N}$ denote the layer input of $N$ token columns, optionally augmented with prompt tokens $\bm{p} \in \{\bm{g}, \bm{e}_t\}$ from \Cref{sub_sec:coupling}; the un-prompted case is recovered by $\bm{z} = \bm{x}$. The masked projections are
\begin{equation}\label{eq:soft_qkv}
\bm{q} = (\bm{\theta}^Q \odot \bm{m}_t^Q)\, \bm{z},\quad
\bm{k} = (\bm{\theta}^K \odot \bm{m}_t^K)\, \bm{z},\quad
\bm{v} = (\bm{\theta}^V \odot \bm{m}_t^V)\, \bm{z},
\end{equation}
and, with $\bm{q}_i, \bm{k}_i, \bm{v}_i$ denoting the $i$-th of $s$ heads, each head computes scaled dot-product attention
\begin{equation}\label{eq:attn}
\mathrm{head}_i
=
\bm{v}_i\,
\mathrm{softmax}\!\left(
    \frac{\bm{q}_i^{\top} \bm{k}_i}{\sqrt{D/s}}
\right),
\end{equation}
so that the soft-MSA output is
\begin{equation}\label{eq:soft_msa}
\mathrm{Soft\text{-}MSA}_t(\bm{z})
=
(\bm{\theta}^O \odot \bm{m}_t^O)\,
\big[\mathrm{head}_1; \cdots; \mathrm{head}_s\big],
\end{equation}
where $[\cdot\,;\cdot]$ concatenates head outputs along the feature dimension. Masks are applied to a selected subset $\mathcal{L}_m$ of MSA layers---for example, the last three layers of ViT-B/16; all other computation is that of the frozen backbone.

\noindent
\textbf{How soft masks find task-specific subnetworks.}
Because $\bm{m}_t^{(1)} = \mathbf{1}$, the initial computation coincides exactly with the pre-trained model. Unlike hard-masking methods that require an explicit selection mechanism such as importance scores, Fisher information, or straight-through estimators~\cite{mallya2018piggyback,Serra2018,wortsman2020supermasks,kang2022forget}, Soft-TF needs none: the selection is performed implicitly by the geometry of the mask parameterization. By the chain rule applied to \Cref{eq:effective}, $\nabla_{\bm{m}} \mathcal{L} = \bm{\theta} \odot \nabla_{\bm{w}} \mathcal{L}$, so one SGD step in mask space is equivalent to a \emph{rescaled} step in weight space:
\begin{equation}\label{eq:precond}
\bm{w}_t^{(k+1)}
=
\bm{w}_t^{(k)} - \eta\, \bm{\theta}^{\odot 2} \odot \nabla_{\bm{w}} \mathcal{L}\big(\bm{w}_t^{(k)}\big),
\end{equation}
i.e., gradient descent preconditioned by $\mathrm{diag}(\bm{\theta}^{\odot 2})$. Synapses with large pre-trained magnitude—those constituting the dominant, well-trained pathways of the backbone—receive proportionally larger updates, while near-zero synapses remain effectively frozen. Optimization is thus intrinsically biased toward modulating the strong pre-trained pathways, which is precisely the soft, optimization-driven subnetwork selection that WLTH postulates; no auxiliary scoring machinery is required. Empirically, the learned masks stay concentrated around $1$ in \Cref{fig:soft_weight_cifar10}, confirming that Soft-TF gently re-weights existing pathways rather than re-learning weights.

\subsection{Coupling Soft Subnetworks with Dual Prompts}\label{sub_sec:coupling}
Following DualPrompt~\cite{wang2022dualprompt}, Soft-TF employs a shared G-prompt $\bm{g} = \{ \bm{g}^{l} \}_{l = l_g^{\text{start}}}^{l_g^{\text{end}}}$ and per-task E-prompts $\bm{e}_t = \{ \bm{e}_t^{l} \}_{l = l_e^{\text{start}}}^{l_e^{\text{end}}}$ attached to disjoint layer ranges, $[l_g^{\text{start}}, l_g^{\text{end}}] \cap [l_e^{\text{start}}, l_e^{\text{end}}] = \emptyset$. Prompts and masks play complementary roles: G-prompts impart task-invariant cues at the input of each attention layer, while E-prompts and the soft masks $\bm{m}_t$ provide task-specific modulation—at the input level and the weight level, respectively. As shown in \Cref{fig:dual_soft}, the E-prompt and the mask set of a task are always selected \emph{jointly}, by the shared key-matching rule of \Cref{eq:key_match}, so Soft-TF adds no separate routing module. We emphasize that Soft-TF is a plug-in: the same construction applies unchanged to L2P with pool prompts and to NoRGa with gated prompts, which we evaluate in \Cref{sec:exp}.

\subsection{Training Objective}\label{sub_sec:objective}
For task $t$, the predictive model is $f_{t}(\bm{x}) = f_{\phi}\big( f_{\bm{\theta}, \bm{g}, \bm{e}_t, \bm{m}_t}(\bm{x}) \big)$, where $f_{\phi}$ is a lightweight classifier head. Training minimizes
\begin{equation}\label{eq:objective}
\min_{\bm{g},\, \bm{e}_t,\, \bm{m}_t,\, \bm{k}_t,\, \phi}
    \;
    \mathcal{L}_{\mathrm{CE}}
    \big(
        f_{\phi}\big(f_{\bm{\theta},\bm{g},\bm{e}_t,\bm{m}_t}(\bm{x})\big),\, y
    \big)
    +
    \lambda\,
    \gamma\big(q(\bm{x}), \bm{k}_t\big),
\end{equation}
where $\mathcal{L}_{\mathrm{CE}}$ is the cross-entropy loss, $\bm{k}_t$ is the task-level key, and $\gamma$ is the cosine-distance regularizer of \Cref{eq:key_match} that pulls queries of task-$t$ samples toward $\bm{k}_t$ to enable test-time task inference. Only the prompts, masks, keys, and classifier are updated; the backbone $\bm{\theta}$ remains frozen. The task index $t$ in \Cref{eq:objective} is available \emph{during training only}, as in the standard CIL protocol; test-time behavior is described next. The full training and test procedures are given in \Cref{alg:algo_train} and \Cref{alg:algo_test}.

\begin{algorithm}[ht]
    \caption{Training Procedure of DualPrompt-Soft-TF}\label{alg:algo_train}
    \small
    \begin{algorithmic}[1]
    \STATE \textbf{Input}: Pre-trained transformer backbone $f$, classifier $f_\phi$;
    \STATE ~~~~ number of tasks $\mathcal{T}$; training sets $\{\{(\bm{x}_{i,t}, y_{i,t})\}_{i=1}^{n_t}\}_{t=1}^{\mathcal{T}}$;
    \STATE ~~~~ G-Prompt $\bm{g}$, E-Prompts $\bm{E}=\{\bm{e}_t\}_{t=1}^{\mathcal{T}}$; 
    \STATE ~~~~ task keys $\bm{K}=\{\bm{k}_t\}_{t=1}^{\mathcal{T}}$;
    \STATE ~~~~ soft-networks $\bm{M}=\{\bm{m}_t\}_{t=1}^{\mathcal{T}}$;
    \STATE ~~~~ prompt insertion layers $[l^{start}_g,l^{end}_g]$, $[l^{start}_e,l^{end}_e]$;
    \STATE ~~~~ prompting function $f_{\bm{\theta}\odot\bm{m}}$; epochs $\mathcal{K}_t$ for task $t$.
    \STATE \textbf{Initialize}: $\phi$, $\bm{g}$, $\bm{E}$, $\bm{M}$, $\bm{K}$
    \FOR{task $t = 1,\dots,\mathcal{T}$}
        \STATE Select task-specific $\bm{e}_t$, soft-network $\bm{m}_t$, and key $\bm{k}_t$
        \STATE Construct prompted backbone $f_{\bm{g},\bm{e}_t,\bm{m}_t}$ by attaching:
        \STATE ~~~~ G-Prompt to layers $[l^{start}_g,l^{end}_g]$,
        \STATE ~~~~ E-Prompt to soft-MSA layers $[l^{start}_e,l^{end}_e]$ via $f_{\bm{\theta}\odot\bm{m}_t}$
        \FOR{epoch $s = 1,\dots,\mathcal{K}_t$}
            \STATE Sample mini-batch $B=\{(\bm{x}_{i,t}, y_{i,t})\}_{i=1}^l$
            \FOR{each $(\bm{x},y)$ in $B$}
                \STATE Compute prompted representation:
                \STATE ~~~~ $\bm{h} = f_{\bm{g},\bm{e}_t,\bm{m}_t}(\bm{x})$
                \STATE Compute prediction $\hat{y}=f_\phi(\bm{h})$
                \STATE Compute per-sample loss:
                \STATE ~~~~ $\mathcal{L}_x = \mathcal{L}_{CE}(\hat{y},y) + \lambda \,\gamma(q(\bm{x}), \bm{k}_t)$
            \ENDFOR
            \STATE Update $\phi$, $\bm{g}$, $\bm{E}$, $\bm{M}$, $\bm{K}$ via backpropagation
        \ENDFOR
    \ENDFOR
    \end{algorithmic}
\end{algorithm}

\begin{algorithm}[ht]
    \caption{DualPrompt-Soft-TF at Test Time}\label{alg:algo_test}
    \small
    \begin{algorithmic}[1]
    \REQUIRE frozen backbone $f_{\bm{\theta}}$; classifier $f_\phi$; G-Prompt $\bm{g}$;
             E-Prompts $\{\bm{e}_t\}_{t=1}^{\mathcal{T}}$; task keys $\{\bm{k}_t\}_{t=1}^{\mathcal{T}}$;
             soft-networks $\{\bm{m}_t\}_{t=1}^{\mathcal{T}}$;
             prompt layers $[l^{start}_g, l^{end}_g]$, $[l^{start}_e, l^{end}_e]$;
             test example $\bm{x}$
    \ENSURE prediction $\hat{y}$
    \IF{task inference by \textit{Prompt ID}}
        \STATE $\hat{t} \gets \arg\min_{t}\, \gamma\big(q(\bm{x}), \bm{k}_t\big)$
               \COMMENT{query feature $q(\bm{x})$}
    \ELSIF{task inference by \textit{Gradient ID}}
        \STATE $\alpha_t \gets 1/\mathcal{T}$ for $t = 1, \dots, \mathcal{T}$
        \STATE $\bm{m}_{\alpha} \gets \sum_{t=1}^{\mathcal{T}} \alpha_t \bm{m}_t$
               \COMMENT{soft-merged network}
        \STATE $\mathcal{H}_{\alpha} \gets \mathcal{H}\big(f_{\bm{\theta}\odot \bm{m}_{\alpha}}(\bm{x})\big)$
               \COMMENT{entropy of mixed model}
        \STATE $\hat{t} \gets \arg\min_{t}\, \partial \mathcal{H}_{\alpha} / \partial \alpha_t$
    \ENDIF
    \STATE Select E-Prompt $\bm{e}_{\hat{t}}$ and soft-network $\bm{m}_{\hat{t}}$
    \STATE Construct $f_{\bm{g},\bm{e}_{\hat{t}},\bm{m}_{\hat{t}}}$: attach $\bm{g}$ at
           $[l^{start}_g, l^{end}_g]$ and $\bm{e}_{\hat{t}}$ at soft-MSA layers
           $[l^{start}_e, l^{end}_e]$ via $f_{\bm{\theta}\odot\bm{m}_{\hat{t}}}$
    \STATE \textbf{return} $\hat{y} = f_\phi\big(f_{\bm{g},\bm{e}_{\hat{t}},\bm{m}_{\hat{t}}}(\bm{x})\big)$
    \end{algorithmic}
\end{algorithm}

\subsection{Test-Time Task Inference}\label{sub_sec:task_inference}
At inference, no task identity is provided. As a plug-in, Soft-TF adopts its host's selector; we use three:
\begin{itemize}[leftmargin=*]
    \item \textbf{Prompt ID} selects the nearest task key, $\hat{t} = \operatorname*{argmin}_t \gamma(q(\bm{x}), \bm{k}_t)$, from \Cref{eq:key_match}; native to L2P and DualPrompt at no extra cost.
    \item \textbf{Gradient ID} weights the masks by $\alpha_t$ with $\sum_t\alpha_t=1$ and infers $\hat{t} = \operatorname*{argmin}_t \partial \mathcal{H}(f_{\bm{\theta}\odot(\sum_t \alpha_t \bm{m}_t)}(\bm{x})) / \partial \alpha_t$ in one forward-backward pass: the mask that most reduces predictive entropy.
    \item \textbf{TII}, native to HiDe-Prompt and NoRGa~\cite{wang2023HiDe,le2025NoRGa}, is a learned selector. It fits a Gaussian $\mathcal{N}(\bm{\mu}_c,\bm{\Sigma}_c)$ per class over the \emph{uninstructed} features $q(\bm{x})$, trains a rehearsal-free all-class classifier by sampling from these Gaussians, and maps its prediction to a task index; a new task only appends its statistics.
\end{itemize}
Given the correct index, the stored $(\bm{e}_t, \bm{m}_t)$ are never overwritten, so the CIL error splits into a within-task error and a task-inference error, the latter dominating as $\mathcal{T}$ grows. We quantify both in \Cref{sec:exp}, and treat the selector as an orthogonal axis shared by all modular CL methods from which Soft-TF benefits directly.

\subsection{Parameter and Inference Cost}\label{sub_sec:cost}
Masking the four $D \times D$ projections of a layer, where $D$ is the model width ($D=768$ for ViT-B/16), costs $4D^2$ parameters, so a task adds $4D^2 |\mathcal{L}_m|$ mask parameters over the masked layer set $\mathcal{L}_m$---about $2.3$M per ViT-B/16 layer and roughly $7$M per task when masking the last three layers, i.e., the $6.93$M budget reported in \Cref{tab:main_vit_backbone}. This is larger than prompt-only budgets, and we therefore report trainable-parameter counts alongside accuracy for all methods and include Adapter and LoRA baselines at comparable budgets in \Cref{sec:exp}. Because masks are folded into the frozen weights once per selected task---$\bm{w}_t = \bm{\theta} \odot \bm{m}_t$ can be precomputed or applied on the fly---the \emph{inference} architecture and FLOPs are identical to the unmodified backbone—no extra modules, tokens, or routing layers. Mask storage grows linearly in $\mathcal{T}$, which we discuss in \Cref{sec:conclusion}.

\subsection{Stability and Bounded Parameter Drift}\label{sec:stability}
We analyze mask-space optimization to explain why Soft-TF adapts while drifting little from the pre-trained solution. The analysis is a local one: we adopt convexity and Lipschitz assumptions that are standard in optimization-based analyses of fine-tuning but do not hold globally for deep networks; the value of the result is the \emph{dependence} of the bound on the initialization quality.

\begin{theorem}[informal; see \Cref{app:convergence}]\label{thm:drift}
Let $g(\bm{m}) = f(\bm{\theta} \odot \bm{m})$ denote the task loss in mask coordinates, assume $g$ is convex with $\|\nabla g\| \le \rho$ on the region of interest, and let $\bm{m}^{\ast}$ be a task-optimal mask with $B = \|\mathbf{1} - \bm{m}^{\ast}\|$. After $K$ mask-space SGD steps started at $\bm{m}^{(1)} = \mathbf{1}$ with $\eta = B / (\rho\sqrt{K})$, the averaged iterate $\bar{\bm{m}} = \frac{1}{K}\sum_k \bm{m}^{(k)}$ satisfies
\begin{equation}\label{eq:regret}
g(\bar{\bm{m}}) - g(\bm{m}^{\ast})
\;\le\;
\frac{B\,\rho}{\sqrt{K}},
\end{equation}
and every effective-weight iterate $\bm{w}^{(k)} = \bm{\theta} \odot \bm{m}^{(k)}$ remains within distance $(1+\sqrt{2})\,\|\bm{\theta}\|_\infty\, B$ of the pre-trained weights $\bm{\theta}$.
\end{theorem}

The bound is governed by $B = \|\mathbf{1} - \bm{m}^{\ast}\|$: how far a task-optimal soft configuration lies from the identity mask, i.e., from the pre-trained solution itself. WLTH is exactly the assertion that $B$ is small for a well-pre-trained backbone. Under this reading, \Cref{thm:drift} implies that mask-only fine-tuning stays confined to a narrow neighborhood of the pre-trained solution manifold and needs fewer steps to reach a given accuracy when $B$ is small. The bounded-drift prediction is corroborated directly---learned masks concentrate around $\mathbf{1}$ in \Cref{fig:soft_weight_cifar10}---and its dependence on initialization quality matches the stronger-pre-training gains in \Cref{tab:main_WLTH}, though we do not measure optimization speed directly. We stress that bounded drift here is a \emph{designed consequence} of freezing $\bm{\theta}$ and parameterizing updates through $\bm{m}_t$—the theorem formalizes the mechanism rather than claiming that the constraint alone eliminates forgetting.

\noindent
\textbf{Two consequences.} Bounded drift yields the two properties Soft-TF relies on, formalized in \Cref{app:convergence}. First, since $\bm{\theta}$ is frozen and each $\bm{m}_t$ is stored and never overwritten, a task's function is time-invariant and, given correct routing, forgetting is structural rather than optimization-dependent, as stated in \Cref{prop:forgetfree}. Second, because every task subnetwork lies in the same $\mathcal{O}(\|\bm{\theta}\|_\infty B)$ ball around $\bm{\theta}$, any two masks are close and a mis-selected mask perturbs the output by only $\mathcal{O}(\|\bm{\theta}\|_\infty B)$, leaving it near the generalist backbone, as shown in \Cref{cor:app_proximity}. The wrong-mask prediction thus stays resolvable by the shared classifier, so class-incremental accuracy decouples from task-inference accuracy---the robustness we verify in \Cref{sec:exp}.

\section{Experiments}\label{sec:exp}
We validate our Soft-TF\footnote{Code is available at \url{https://github.com/ihaeyong/Soft-TF}.} on several benchmark datasets against continuous learning baselines in Class-Incremental Learning (CIL).

\subsection{Experimental Settings}

\paragraph{Datasets.} We evaluate our method mainly on 10/20-Split-CIFAR100~\citep{krizhevsky2009learning}, constructed by splitting the 100 classes into 10 or 20 tasks, and 10-Split-ImageNet-R~\citep{hendrycks2021many}, constructed by splitting the 200 classes into 10 tasks. To show our effectiveness, we additionally compare our method with the baselines on 5-Split-CUB200 and 5-Datasets.


\noindent
\textbf{Baselines.}
We compare against four families. \emph{Prompt-based}: L2P~\citep{wang2022learning}, DualPrompt~\citep{wang2022dualprompt}, CODA-Prompt~\cite{Smith_2023_CVPR}, S-Prompt~\cite{wang2022s}, HiDe-Prompt~\cite{wang2023HiDe}, ESN~\cite{wang2022esn}, CPrompt~\cite{gao2024cprompt}, and NoRGa~\cite{le2025NoRGa}. \emph{Task-specific PEFT}: Adapter~\citep{houlsby2019parameter} and LoRA~\citep{hu2021lora}. \emph{Rehearsal-free}: RanPAC~\citep{mcdonnell2023ranpac}. And joint-training upper bounds. RanPAC is a complementary contrast---it needs no task inference but performs no per-task backbone specialization---so we run both its analytic prototype head and its full Phase-1 PETL variant to separate the head from backbone adaptation. All reproductions, denoted $\ast$, use the official code under our exact backbone, class order, and splits. The Task ID column of \Cref{tab:main_cil_time} names each method's native selector: beyond the three mechanisms of \Cref{sub_sec:task_inference}, ``KNN'' is S-Prompt's nearest-centroid matching, and ``None'' marks no explicit task selection, as in CODA-Prompt's shared-component attention and RanPAC's all-class head. Soft-TF applies only to hosts whose selector yields a discrete task index, so CODA-Prompt is excluded by design, with continuous mask composition left as future work. Soft-TF inherits its host's selector---Prompt ID for L2P and DualPrompt, learned TII for HiDe-Prompt and NoRGa, Gradient ID where noted; the HiDe-Prompt + Soft-TF row uses the official HiDe/NoRGa framework, with the controlled comparison in \Cref{tab:task_id_acc} and \Cref{fig:softtf_matrix}.

\noindent
\textbf{Trainable-parameter fairness.}
Soft-TF's per-task mask budget is larger than a prompt-only budget, as quantified in \Cref{sub_sec:cost}, so accuracy comparisons against prompt-only baselines alone would be uninformative about where the gains come from. We therefore report trainable-parameter counts alongside accuracy in \Cref{tab:main_cil_time,tab:main_vit_backbone}, include Adapter and LoRA baselines whose trainable budgets are comparable to or exceed Soft-TF's, and provide layer-restricted variants of Soft-TF that mask only one to three attention layers and shrink the budget while retaining most of the benefit, as shown in \Cref{fig:layer_soft_cifar10}. Across these matched-budget comparisons, the advantage of multiplicative, well-initialized masks persists, indicating that the gains are attributable to the parameterization rather than to raw capacity.

\begin{table*}[!ht]
\begin{center}
\caption{\small \textbf{Vision class-incremental learning} on 10/20-Split-CIFAR100 and 10-Split-ImageNet-R: accuracy, forgetting, trainable parameters, and train/test time. $\ast$ marks our reproductions; L[10,11,12] denotes Soft-TF on the last three attention layers.}

\resizebox{0.99\textwidth}{!}{
\begin{tabular}{l c c *{6}{r!{\,/\,}l}}
\toprule
& \textbf{ViT-B/16} & & \multicolumn{4}{c}{\textbf{10-Split-CIFAR100}} & \multicolumn{4}{c}{\textbf{20-Split-CIFAR100}} & \multicolumn{4}{c}{\textbf{10-Split-ImageNet-R}} \\
\cmidrule(lr){4-7}\cmidrule(lr){8-11}\cmidrule(lr){12-15}
\multicolumn{1}{c}{\textbf{Method}} & \multicolumn{1}{c}{\textbf{\#Tr.Params.}} & \textbf{Task ID} & \multicolumn{2}{c}{\textbf{ACC/Forget}} & \multicolumn{2}{c}{\textbf{Tr./Test[sec]}} & \multicolumn{2}{c}{\textbf{ACC/Forget}} & \multicolumn{2}{c}{\textbf{Tr./Test[sec]}} & \multicolumn{2}{c}{\textbf{ACC/Forget}} & \multicolumn{2}{c}{\textbf{Tr./Test[sec]}} \\ \midrule

L2P$^\ast$           & 0.03M & Prompt ID & 83.77 & 6.63 & 12.00K & 75 & 71.29 & 13.96 & 11.50K & 76 & 60.44 & 9.00 & 12.80K & 46 \\
DualPrompt$^{\ast}$  & 0.03M & Prompt ID & 86.50 & 5.77 & 12.12K & 76 & 82.98 & 8.20  & 11.60K & 78 & 68.13 & 4.46 & 13.10K & 47 \\
S-Prompt$^{\ast}$    & 0.03M & KNN       & 87.57 & 3.63 & 12.12K & 76 & 84.90 & 7.05  & 11.60K & 78 & 74.25 & 4.73 & 13.10K & 47 \\
CODA-Prompt$^{\ast}$ & 0.03M & None      & 86.94 & 4.04 & 12.12K & 76 & 84.70 & 7.11  & 11.60K & 78 & 74.26 & 5.17 & 13.10K & 47 \\
HiDe-Prompt$^{\ast}$ & 0.03M & TII       & 94.57 & 1.59 & 12.12K & 76 & 94.66 & 1.42  & 11.60K & 78 & 75.08 & 4.23 & 13.10K & 47 \\
NoRGa$^{\ast}$       & 0.03M & TII       & 94.71 & 1.59 & 12.12K & 76 & 94.66 & 1.73  & 11.60K & 78 & 74.60 & 4.67 & 13.10K & 47 \\
\midrule

DualPrompt + Adapter & 6.93M & Prompt ID & 86.51 & 4.75 & 14.98K & 90 & 84.48 & 5.81 & 16.13K & 109 & 70.56 & 4.71  & 15.65K & 54 \\
DualPrompt + LoRA    & 6.93M & Prompt ID & 82.00 & 4.33 & 13.24K & 79 & 92.14 & 2.02 & 15.89K & 105 & 43.51 & 13.20 & 15.09K & 53 \\
\midrule

RanPAC$^\ast$ (RP head)       & 0.2M  & None & 88.97 & 4.41 & 0.83K & 91  & 89.02 & 4.34 & 1.15K & 78 & 67.24 & 5.14 & 0.60K & 82 \\
~+ Phase-1 PETL (full RanPAC) & 1.20M & None & 91.79 & 3.30 & 2.37K & 135 & 90.99 & 3.69 & 1.67K & 50 & 74.99 & 4.26 & 1.47K & 74 \\
\midrule

L2P + \textbf{Soft-TF}-L[3,4,5]           & 6.93M & Prompt ID & 86.26 & 4.79 & 12.85K & 78 & 76.17 & 15.77 & 13.98K & 100 & 69.80 & 5.13 & 14.23K & 49 \\
DualPrompt + \textbf{Soft-TF}-L[10,11,12] & 6.93M & Prompt ID & 92.35 & 2.98 & 13.87K & 80 & 97.40 & 0.57  & 15.60K & 104 & 76.62 & 5.30 & 15.35K & 52 \\
NoRGa + \textbf{Soft-TF}-L[10,11,12]       & 6.93M & TII & 98.38 & 0.09 & 13.87K & 80 & 98.74 & 0.08 & 15.60K & 104 & 86.71 & 0.92 & 15.35K & 52 \\
\rowcolor{gg} HiDe-Prompt + \textbf{Soft-TF}-L[10,11,12] & 6.93M & TII & \textbf{98.39} & \textbf{0.04} & 13.87K & 80 & \textbf{98.88} & \textbf{0.07} & 15.60K & 104 & \textbf{86.93} & \textbf{0.92} & 15.35K & 52 \\
\midrule

Joint of DualPrompt          & 0.13M & - & \multicolumn{2}{c}{90.85} & \multicolumn{2}{c}{-} & \multicolumn{2}{c}{90.85} & \multicolumn{2}{c}{-} & \multicolumn{2}{c}{79.13} & \multicolumn{2}{c}{-} \\
Joint of DualPrompt+Soft-TF  & 6.93M & - & \multicolumn{2}{c}{95.90} & \multicolumn{2}{c}{-} & \multicolumn{2}{c}{95.90} & \multicolumn{2}{c}{-} & \multicolumn{2}{c}{80.21} & \multicolumn{2}{c}{-} \\
Joint of NoRGa+Soft-TF       & 6.93M & - & \multicolumn{2}{c}{98.20} & \multicolumn{2}{c}{-} & \multicolumn{2}{c}{98.20} & \multicolumn{2}{c}{-} & \multicolumn{2}{c}{90.73} & \multicolumn{2}{c}{-} \\
\bottomrule
\end{tabular}}
\label{tab:main_cil_time}
\end{center}
\end{table*}

\begin{figure*}[!t]
    \centering
    \small
    \setlength{\tabcolsep}{6pt}{%
    \begin{tabular}{cc}
    \includegraphics[width=0.44\textwidth]{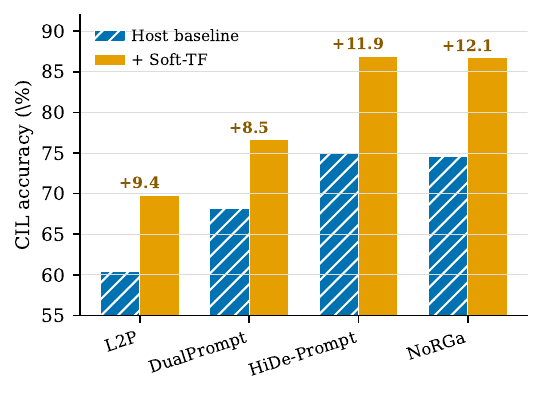} &
    \includegraphics[width=0.46\textwidth]{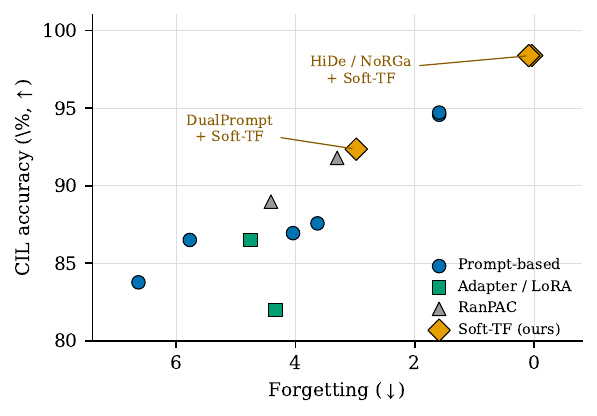} \\
    \small (a) Plug-in gain of Soft-TF across four hosts (10-Split-ImageNet-R) &
    \small (b) Accuracy--forgetting trade-off (10-Split-CIFAR100) \\
    \end{tabular}
    }
    \caption{\small \textbf{Summary of \Cref{tab:main_cil_time}.} (a) Adding Soft-TF lifts every prompt-based host by $+8.5$ to $+12.1$\%p, confirming it is a general plug-in rather than a standalone method. (b) Across all compared methods, Soft-TF variants occupy the top-right frontier---highest accuracy at the lowest forgetting---dominating prompt-based, adapter/LoRA, and RanPAC baselines.}
    \label{fig:table1_summary}
    \vspace{-0.1in}
\end{figure*}





\begin{table}[ht]

\begin{center}
\caption{\small \textbf{DualPrompt + Soft-TF across ViT backbones} on 10-Split-ImageNet-R. Larger backbones raise accuracy and lower forgetting at a higher parameter and compute cost.}

\resizebox{0.48\textwidth}{!}{
\begin{tabular}{l r r r r!{\,/\,}l}
\toprule
\textbf{DualPrompt} & & \multicolumn{4}{c}{\textbf{10-Split-ImageNet-R}} \\
\cmidrule(lr){3-6}
\textbf{~+Soft-TF} & \textbf{\#Tr.Prams} & \textbf{ACC($\uparrow$)} & \textbf{Forget($\downarrow$)} & \multicolumn{2}{c}{\textbf{Tr./Test[sec]}} \\ \midrule
ViT-Ti/16 (5.7M) & 0.41M  & 49.34          & 9.02          & 15.10K & 50 \\
ViT-S/16 (22M)   & 1.71M  & 65.12          & 6.67          & 15.22K & 51 \\
ViT-B/16 (86M)   & 6.93M  & 76.62          & 5.30          & 15.35K & 52 \\
\rowcolor{gg} ViT-L/16 (307M)  & 12.50M & \textbf{78.86} & \textbf{4.46} & 28.57K & 99 \\
\bottomrule
\end{tabular}}

\label{tab:main_vit_backbone}
\end{center}
\vspace{-0.15in}
\end{table}

\subsection{Performances}
\textbf{Performances of Soft-TF.}
We compare our Soft-TF with state-of-the-art CIL baselines, as shown in \Cref{tab:main_cil_time} and summarized in \Cref{fig:table1_summary}. Soft-TF consistently outperforms prompt-based and parameter-efficient fine-tuning methods—including L2P, DualPrompt, NoRGa, Adapter, and LoRA—in both accuracy and forgetting. In particular, NoRGa+Soft-TF improves NoRGa by up to $+12.1$\%p (74.60 to 86.71 on 10-Split-ImageNet-R); NoRGa mitigates task interference but adapts only at the input, through global or head-level prompt modulation. By contrast, Soft-TF directly modulates attention projection matrices through task-specific soft subnetworks, enabling fine-grained and continuous adaptation within a frozen Transformer backbone, which results in higher accuracy and significantly reduced forgetting across all benchmarks. At a trainable-parameter budget comparable to the Adapter and LoRA variants, Soft-TF achieves these gains while keeping the inference cost identical to the unmodified backbone. Remarkably, Soft-TF can even surpass the \emph{joint-training upper bound of DualPrompt}: NoRGa+Soft-TF exceeds it on all three benchmarks and DualPrompt+Soft-TF exceeds it on both CIFAR100 splits, as it explicitly identifies and optimizes distinct task-specific soft subnetworks for each incremental task, achieving near-optimal per-task specialization, whereas joint training is constrained to a single shared solution that must compromise across tasks. We note that the joint upper bound is an upper bound only for methods that share one parameter set across all classes; a system that stores per-task parameters and selects them with sufficiently accurate task inference is not bounded by it, which is precisely the regime in which Soft-TF operates.

\noindent
\textbf{Comparison with Prompt-based State of the Art.}
\Cref{tab:main_cil_time} also includes reproduced prompt-based state-of-the-art baselines—S-Prompt, CODA-Prompt, and HiDe-Prompt—under the same protocol. While improved prompt representations generally lead to better continual learning performance, their adaptation capacity remains bounded by prompt-level input conditioning; Soft-TF, which additionally modulates the attention projections through task-specific soft subnetworks, consistently achieves the best accuracy--forgetting trade-off---NoRGa+Soft-TF reaches 98.38\% versus 94.71\% for the strongest prompt baseline, NoRGa, on 10-Split-CIFAR100. This addresses the expressiveness gap directly rather than through richer prompt engineering.


\noindent
\textbf{Comparison with RanPAC.}
\Cref{tab:main_cil_time} also reports RanPAC under our controlled protocol. Even its full variant with Phase-1 PETL trails DualPrompt+Soft-TF on every benchmark---91.79 versus 92.35 on 10-Split-CIFAR100, 90.99 versus 97.40 on 20-Split-CIFAR100, and 74.99 versus 76.62 on 10-Split-ImageNet-R---and trails NoRGa+Soft-TF by margins of up to $+12$\%p on ImageNet-R. Notably, the gap widens as the number of tasks grows: from 10 to 20 splits, RanPAC's single shared prototype head must accommodate more classes without any per-task specialization and stagnates around 91\%, whereas per-task soft subnetworks retain specialization and improve to 98.74\%. Conversely, RanPAC's strengths are also visible in the table—its analytic head needs no task inference and trains at a fraction of the cost—which highlights the complementary design axes discussed in \Cref{sec:related}.

\begin{figure*}[!t]
    \centering
    \small
    \setlength{\tabcolsep}{6pt}{%
    \begin{tabular}{cc}
    \includegraphics[width=0.46\textwidth]{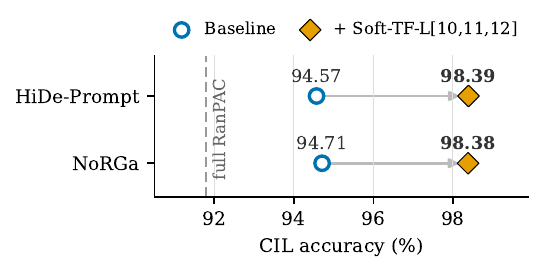} &
    \includegraphics[width=0.46\textwidth]{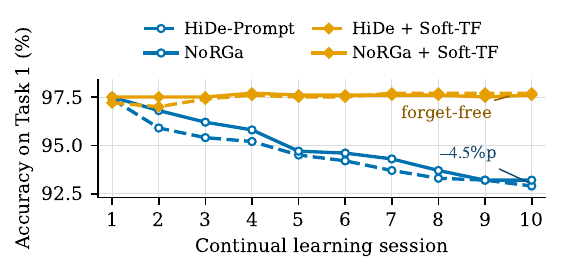} \\
    \small (a) CIL accuracy at fixed task-ID inference with TII $= 81.38$\% &
    \small (b) Accuracy of the most-forgotten task, Task 1, across sessions \\
    \end{tabular}
    }
    \caption{\small \textbf{Effect of Soft-TF under fixed task-ID inference} on 10-Split-CIFAR100, reproduced in the official HiDe/NoRGa framework under our protocol. Both variants of each method share the same learned TII head, so the task-ID inference accuracy is fixed at 81.38\%. In panel (a), adding Soft-TF-L[10,11,12] lifts CIL accuracy by $+3.8$\%p on HiDe-Prompt and $+3.7$\%p on NoRGa; the dashed line marks full RanPAC under the same protocol. In panel (b), Task 1---the task both baselines forget most---decays monotonically by up to $4.5$\%p as later tasks arrive, whereas with Soft-TF the stored task-1 mask is never overwritten and the trajectory stays flat, i.e., forget-free.}
    \label{fig:softtf_matrix}
    \vspace{-0.1in}
\end{figure*}

\noindent
\textbf{Task-Inference Analysis.}
Since per-task masks eliminate parameter interference by construction, as noted in \Cref{sub_sec:task_inference}, the CIL bottleneck is task inference. \Cref{tab:main_app_cil} isolates it for DualPrompt + Soft-TF: switching Prompt ID to Gradient ID raises CIL from 92.35\% to 97.87\% on 10-Split-CIFAR100 and from 76.62\% to 82.38\% on 10-Split-ImageNet-R, so residual error is dominated by selection, not representation. \Cref{tab:task_id_acc} then controls for the selector: reproduced in the official HiDe/NoRGa framework, both variants of a method share the \emph{same} learned TII head, so task-ID accuracy is fixed, yet Soft-TF still raises CIL from 94.71\% to 98.38\% for NoRGa and from 94.57\% to 98.39\% for HiDe-Prompt with forgetting below 0.1. Two conclusions follow. First, these gains come purely from representation specialization, not better selection. Second, a selector wrong on 18.6\% of samples still yields 98.38\% accuracy---a direct consequence of the bounded drift of \Cref{sec:stability}, whereby any two task masks stay near the shared backbone $\bm{\theta}$, so a wrong mask still evaluates a near-generalist function that the shared classifier recovers. This robustness strengthens as tasks get finer and harder, and holds for both hosts alike: on 20-Split-CIFAR100 the shared TII drops to 80.75\% yet Soft-TF lifts HiDe-Prompt by $+4.22$\%p and NoRGa by $+4.08$\%p, both exceeding their $+3.7$--$3.8$\%p at 10 splits, and widening further to $+6.48$\%p for NoRGa on 50-Split-CIFAR100 ($93.04\!\to\!99.52$ at a TII of only $79.62$\%). On 10-Split-ImageNet-R, where TII is correct only 58.24\% of the time, Soft-TF gains $+11.85$\%p for HiDe-Prompt and $+12.11$\%p for NoRGa, cutting forgetting from above 4 to below 1. \Cref{fig:softtf_matrix} visualizes this: at fixed TII, Soft-TF lifts both methods by nearly four points, further widening their lead over full RanPAC, while the most-forgotten task decays up to 4.5\%p under the baselines yet stays flat under Soft-TF.

\begin{table*}[ht]
\begin{center}
\caption{\small \textbf{Prompt ID vs.\ Gradient ID} for DualPrompt + Soft-TF. Both selectors reuse the same trained model, so the accuracy gap isolates the task-inference error.}

\resizebox{0.99\textwidth}{!}{
\begin{tabular}{l c c *{6}{r!{\,/\,}l}}
\toprule
& \textbf{ViT-B/16} & & \multicolumn{4}{c}{\textbf{10-Split-CIFAR100}} & \multicolumn{4}{c}{\textbf{20-Split-CIFAR100}} & \multicolumn{4}{c}{\textbf{10-Split-ImageNet-R}} \\
\cmidrule(lr){4-7}\cmidrule(lr){8-11}\cmidrule(lr){12-15}
\multicolumn{1}{c}{\textbf{Method}} & \multicolumn{1}{c}{\textbf{\#Tr.Params.}} & \textbf{Task ID} & \multicolumn{2}{c}{\textbf{ACC/Forget}} & \multicolumn{2}{c}{\textbf{Tr./Test[sec]}} & \multicolumn{2}{c}{\textbf{ACC/Forget}} & \multicolumn{2}{c}{\textbf{Tr./Test[sec]}} & \multicolumn{2}{c}{\textbf{ACC/Forget}} & \multicolumn{2}{c}{\textbf{Tr./Test[sec]}} \\ \midrule

DualPrompt + \textbf{Soft-TF}-L[3,4,5]    & 6.93M & Prompt ID   & 91.77 & 3.37 & 13.87K & 80  & 94.43 & 2.02 & 15.60K & 104 & 74.70 & 6.46 & 15.35K & 52 \\
DualPrompt + \textbf{Soft-TF}-L[3,4,5]    & 6.93M & Gradient ID & 93.76 & 1.83 & 13.87K & 130 & 95.38 & 1.73 & 15.60K & 163 & 82.15 & 2.20 & 15.35K & 82 \\
\midrule
DualPrompt + \textbf{Soft-TF}-L[10,11,12] & 6.93M & Prompt ID   & 92.35 & 2.98 & 13.87K & 80  & 97.40 & 0.57 & 15.60K & 104 & 76.62 & 5.30 & 15.35K & 52 \\
\rowcolor{gg} DualPrompt + \textbf{Soft-TF}-L[10,11,12] & 6.93M & Gradient ID & \textbf{97.87} & \textbf{0.21} & 13.87K & 130 & \textbf{99.05} & \textbf{0.24} & 15.60K & 163 & \textbf{82.38} & \textbf{0.59} & 15.35K & 82 \\
\bottomrule
\end{tabular}
}
\label{tab:main_app_cil}
\end{center}
\end{table*}

\begin{table}[!t]
\begin{center}
\caption{\small \textbf{Task-ID inference accuracy vs.\ CIL performance} (ViT-B/16, Sup-21K). Each pair shares the same TII head~\cite{le2025NoRGa}, so Soft-TF's CIL gains come purely from representation specialization, not better task selection.}
\resizebox{0.99\columnwidth}{!}{
\begin{tabular}{l c r r r}
\toprule
\textbf{Method} & \textbf{Selector} & \textbf{Task-ID Acc} & \textbf{CIL ACC($\uparrow$)} & \textbf{Forget($\downarrow$)} \\ \midrule
\multicolumn{5}{l}{\itshape 10-Split-CIFAR100} \\
NoRGa$^\ast$                                & TII & 81.38 & 94.71 & 1.59 \\
\rowcolor{gg} ~+ \textbf{Soft-TF}-L[10,11,12]             & TII & 81.38 & \textbf{98.38} & \textbf{0.09} \\ \midrule
HiDe-Prompt$^\ast$                          & TII & 81.38 & 94.57 & 1.59 \\
\rowcolor{gg} ~+ \textbf{Soft-TF}-L[10,11,12]             & TII & 81.38 & \textbf{98.39} & \textbf{0.04} \\
\midrule
\multicolumn{5}{l}{\itshape 20-Split-CIFAR100} \\
NoRGa$^\ast$                                & TII & 80.75 & 94.66 & 1.73 \\
\rowcolor{gg} ~+ \textbf{Soft-TF}-L[10,11,12]             & TII & 80.75 & \textbf{98.74} & \textbf{0.08} \\ \midrule
HiDe-Prompt$^\ast$                          & TII & 80.75 & 94.66 & 1.42 \\
\rowcolor{gg} ~+ \textbf{Soft-TF}-L[10,11,12]             & TII & 80.75 & \textbf{98.88} & \textbf{0.07} \\
\midrule
\multicolumn{5}{l}{\itshape 50-Split-CIFAR100} \\
NoRGa$^\ast$                                & TII & 79.62 & 93.04 & 2.76 \\
\rowcolor{gg} ~+ \textbf{Soft-TF}-L[10,11,12]             & TII & 79.62 & \textbf{99.52} & \textbf{0.11} \\
\midrule
\multicolumn{5}{l}{\itshape 10-Split-ImageNet-R} \\
NoRGa$^\ast$                                & TII & 58.24 & 74.60 & 4.67 \\
\rowcolor{gg} ~+ \textbf{Soft-TF}-L[10,11,12]             & TII & 58.24 & \textbf{86.71} & \textbf{0.92} \\ \midrule
HiDe-Prompt$^\ast$                          & TII & 58.24 & 75.08 & 4.23 \\
\rowcolor{gg} ~+ \textbf{Soft-TF}-L[10,11,12]             & TII & 58.24 & \textbf{86.93} & \textbf{0.92} \\
\bottomrule
\end{tabular}}
\label{tab:task_id_acc}
\end{center}
\vspace{-0.1in}
\end{table}

\subsection{Variants of ViTs.}
\begin{figure}[!ht]
    \centering
    \includegraphics[width=0.48\textwidth]{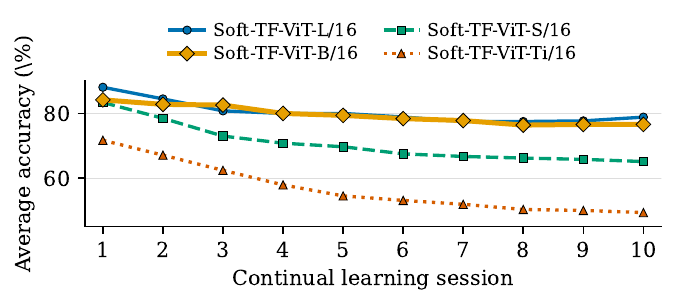}
    \caption{\small \textbf{DualPrompt + Soft-TF across ViT backbones} on 10-Split-ImageNet-R: session-wise average accuracy. Larger backbones degrade more slowly, and ViT-B/16 stays close to ViT-L/16 throughout.}
    \label{fig:main_vit_backbone}
    \vspace{-0.15in}
\end{figure}

Across different ViT backbones, Soft-TF exhibits consistent advantages in both parameter efficiency and continual learning stability on 10-Split-ImageNet-R, as shown in \Cref{tab:main_vit_backbone}. ViT-Ti/16 uses 0.41M trainable parameters and reaches 49.34\% accuracy at 9.02 forgetting; scaling to ViT-S/16, ViT-B/16, and ViT-L/16 raises trainable parameters to 1.71M, 6.93M, and 12.50M and accuracy to 65.12\%, 76.62\%, and 78.86\%, with forgetting falling to 6.67, 5.30, and 4.46. ViT-L/16 gives the best result, 78.86\% at 4.46 forgetting, at about 28K/99\,s train/test time versus 15K/52\,s for ViT-B/16. \Cref{fig:main_vit_backbone} shows larger backbones degrade more slowly across the 10 sessions---ViT-L/16 the slowest---so the accuracy gain trades off against parameter and compute cost.

\subsection{Well-initialized Soft-TF}

\noindent
\textbf{Random initialization.} Random initialization of Soft-Transformer's weights plays a critical role when leveraging well-pretrained models like Vision Transformers (ViTs). The optimal training point is the parameters of a well-pretrained model. Among the initialization methods, Uniform initialization for Soft-TransFormer satisfies this requirement effectively. To test this, we compare Xavier, Kaiming, Normal, and Uniform mask initialization, as shown in \Cref{tab:app_init_perform}. Uniform(1.0,1.0), which starts the masks at the pre-trained solution, reaches 92.35\%/2.98, versus 90.45--90.72\%/3.63--3.85 for Xavier, Kaiming, and Normal, supporting the WLTH view that initializing at the pre-trained point is optimal.

\noindent
\textbf{Well-initialized LTH (WLTH).} 
Based on \Cref{tab:main_WLTH}, we evaluated our Soft-TF on the 10-Split-CIFAR100 and 5-Split-CUB200 datasets using three distinct pre-trained ViT models---ImageNet-21K, SAM, and DINO---to demonstrate its efficacy, including on Well-initialized Lottery Ticket Hypothesis (WLTH) backbones. For the 10-Split-CIFAR100 benchmark, the ImageNet-21K backbone achieved the highest accuracy of 92.35\% and the lowest forgetting rate of 2.98, closely followed by the SAM backbone with 92.30\% accuracy and 2.99 forgetting. Similarly, for the 5-Split-CUB200 benchmark, the ImageNet-21K backbone led with 76.17\% accuracy and 9.04 forgetting, which was also nearly matched by the SAM backbone with 76.11\% accuracy and 9.01 forgetting. In contrast, the self-supervised DINO backbone trails at 84.50\%/12.27 on CIFAR100 and 69.79\%/10.93 on CUB-200. These trends match the bounded-drift analysis of \Cref{sec:stability}: a stronger backbone places a task-optimal mask closer to its identity initialization, a smaller $B$ that the bound ties to less drift and fewer optimization steps.

\begin{table}[ht]
\begin{center}
\caption{\small \textbf{Effect of mask initialization} on DualPrompt + Soft-TF, 10-Split-CIFAR100. Uniform(1.0,1.0), which starts optimization at the pre-trained solution, outperforms Xavier, Kaiming, and Normal initialization.}

\resizebox{0.32\textwidth}{!}{
\begin{tabular}{l r r}
\toprule
 &\multicolumn{2}{c}{\textbf{10-Split-CIFAR100}}   \\
\cmidrule(lr){2-3}
\textbf{Random Init.} & ACC($\uparrow$) & Forget($\downarrow$)  \\ \midrule

 Xavier     & 90.59     &  3.85     \\
 Kaiming    & 90.72     &  3.63     \\
 Normal     & 90.45     & 3.78      \\

 \rowcolor{gg} \textbf{Uniform(1.0, 1.0)}   & \textbf{92.35} &  \textbf{2.98}    \\
\bottomrule
\end{tabular}}
\label{tab:app_init_perform}
\end{center}
\vspace{-0.15in}
\end{table}

\begin{table}[ht]
\begin{center}
\caption{\small \textbf{Effect of the pre-trained backbone} on DualPrompt + Soft-TF, 10-Split-CIFAR100 and 5-Split-CUB200. Stronger ImageNet-21K and SAM backbones outperform the self-supervised DINO backbone.}

\resizebox{0.43\textwidth}{!}{
\begin{tabular}{l r r r r}
\toprule
DualPrompt& \multicolumn{2}{c}{\textbf{10-Split-CIFAR100}} & \multicolumn{2}{c}{\textbf{5-Split-CUB200}}  \\
\cmidrule(lr){2-3}\cmidrule(lr){4-5}
~+Soft-TF &  ACC($\uparrow$) & Forget($\downarrow$) & ACC($\uparrow$) & Forget($\downarrow$)  \\ \midrule
\rowcolor{gg} \textbf{ImageNet-21K}   & \textbf{92.35} & \textbf{2.98}   & \textbf{76.17} & \textbf{9.04}  \\
SAM  & 92.30  & 2.99 & 76.11 & 9.01 \\ 
DINO  & 84.50 & 12.27  & 69.79  & 10.93  \\
\bottomrule
\end{tabular}}
\label{tab:main_WLTH}
\end{center}
\vspace{-0.15in}
\end{table}

\subsection{Soft-MSA Layers}
\begin{figure}[ht]
    \centering
    \small
    \setlength{\tabcolsep}{0pt}{%
    \begin{tabular}{cc}

    \includegraphics[width=0.49\columnwidth]{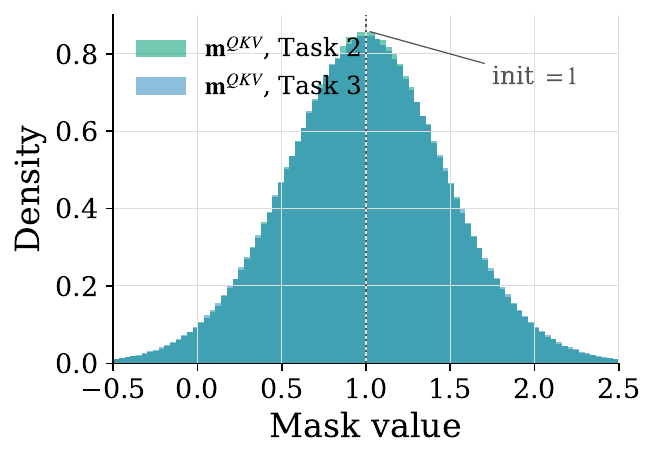} &
    \includegraphics[width=0.49\columnwidth]{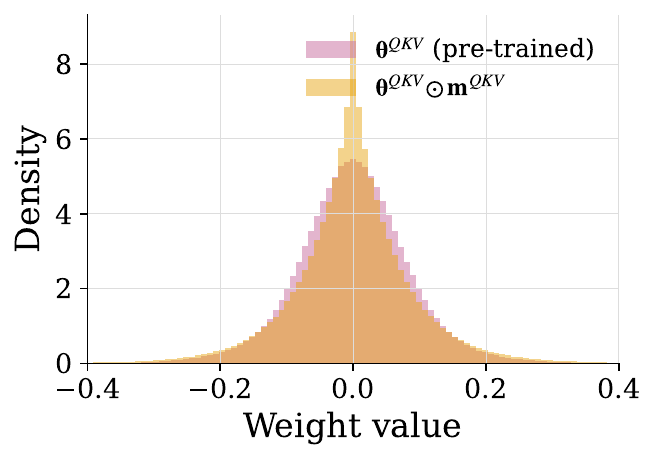} \\
    \small (a) mask $\bm{m}^{QKV}$ & \small (b) $\bm{\theta}^{QKV}$ {v.s.} $\bm{\theta}^{QKV} \odot \bm{m}^{QKV}$ \\
    \end{tabular}
    }
    \caption{\small \textbf{Attention Layer-(L[12]) parameter distributions} of HiDe-Prompt + Soft-TF on 10-Split-CIFAR100. (a) The learned masks $\bm{m}^{QKV}$ stay centered at their initialization of $1$ and are nearly identical across Task 2 and Task 3. (b) The masked weights $\bm{\theta}^{QKV}\odot\bm{m}^{QKV}$ preserve the pre-trained distribution $\bm{\theta}^{QKV}$, remaining centered at $0$ with a comparable spread, so the backbone is only mildly modified.}

    \label{fig:soft_weight_cifar10}
\end{figure}

\Cref{fig:soft_weight_cifar10} shows the histogram density estimates of the attention-layer L[12] parameters for HiDe-Prompt + Soft-TF: the learned masks $\bm{m}^{QKV}$ and the resulting masked weights $\bm{\theta}^{QKV} \odot \bm{m}^{QKV}$. In \Cref{fig:soft_weight_cifar10}(a), the masks remain centered at their initialization value of $\mu \approx 1.0$ with a bell-shaped spread, and the distributions of Task 2 and Task 3 are almost indistinguishable; $\bm{m}^{QKV}$ thus acts as a soft gating mechanism that re-weights, rather than replaces, the pre-trained parameters. In \Cref{fig:soft_weight_cifar10}(b), the pre-trained weights $\bm{\theta}^{QKV}$ are centered at $0$ with a Gaussian-like spread, and the masked weights $\bm{\theta}^{QKV} \odot \bm{m}^{QKV}$ preserve this distribution closely---centered at $0$ with a comparable standard deviation. This heavy overlap confirms that, although the mask itself varies around $1$, the effective weights stay close to the pre-trained backbone, consistent with the bounded-drift analysis of \Cref{sec:stability}: fine-tuning introduces task-specific modulation while retaining the pre-trained structure, which mitigates catastrophic forgetting across sequential tasks.

\begin{figure}[ht]
    \centering
    \small
    \setlength{\tabcolsep}{0pt}{%
    \begin{tabular}{cc}
    \includegraphics[width=0.49\columnwidth, trim={2.6cm 0.78cm 4.5cm 1.2cm},clip]{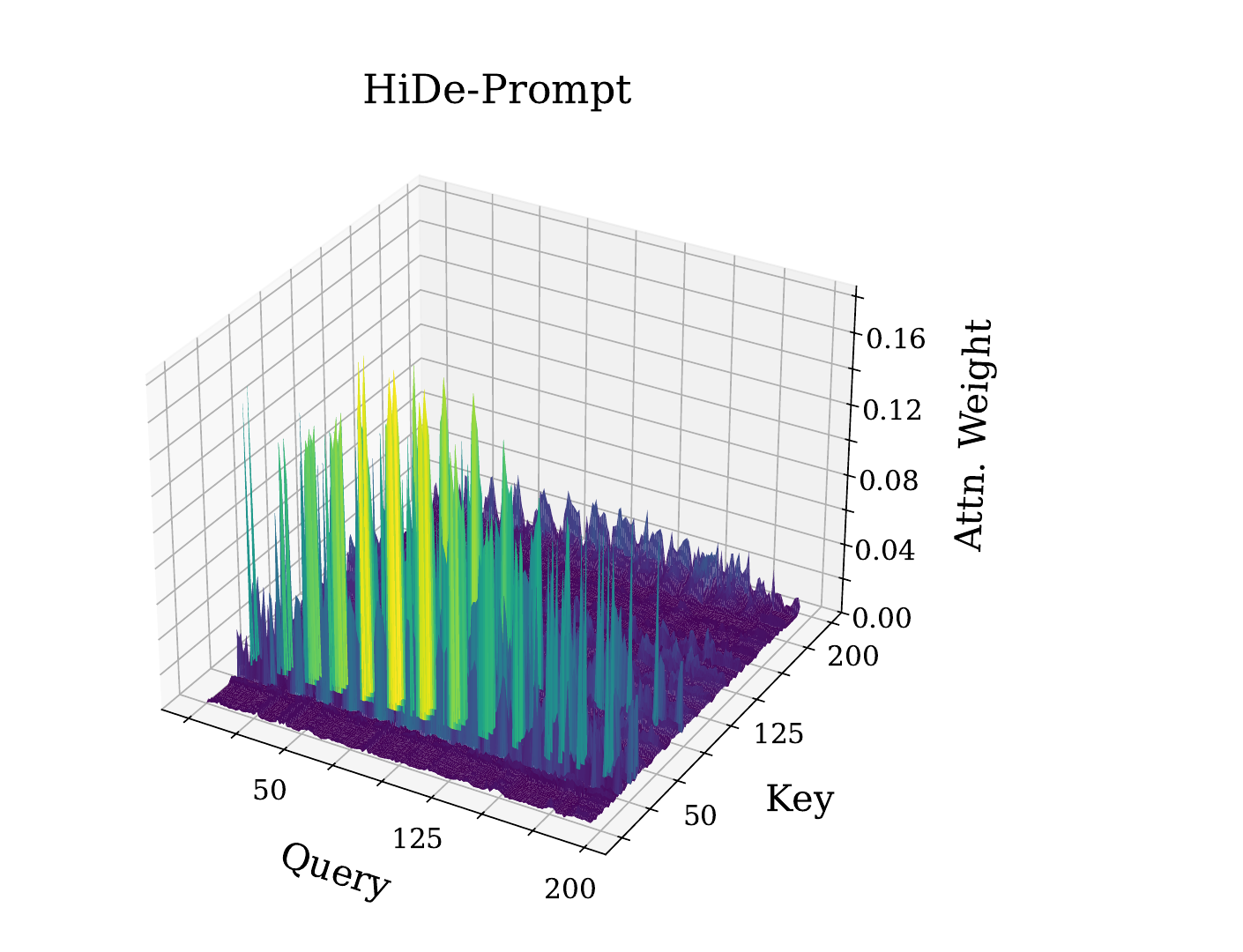} &
    \includegraphics[width=0.49\columnwidth, trim={2.6cm 0.78cm 4.5cm 1.2cm},clip]{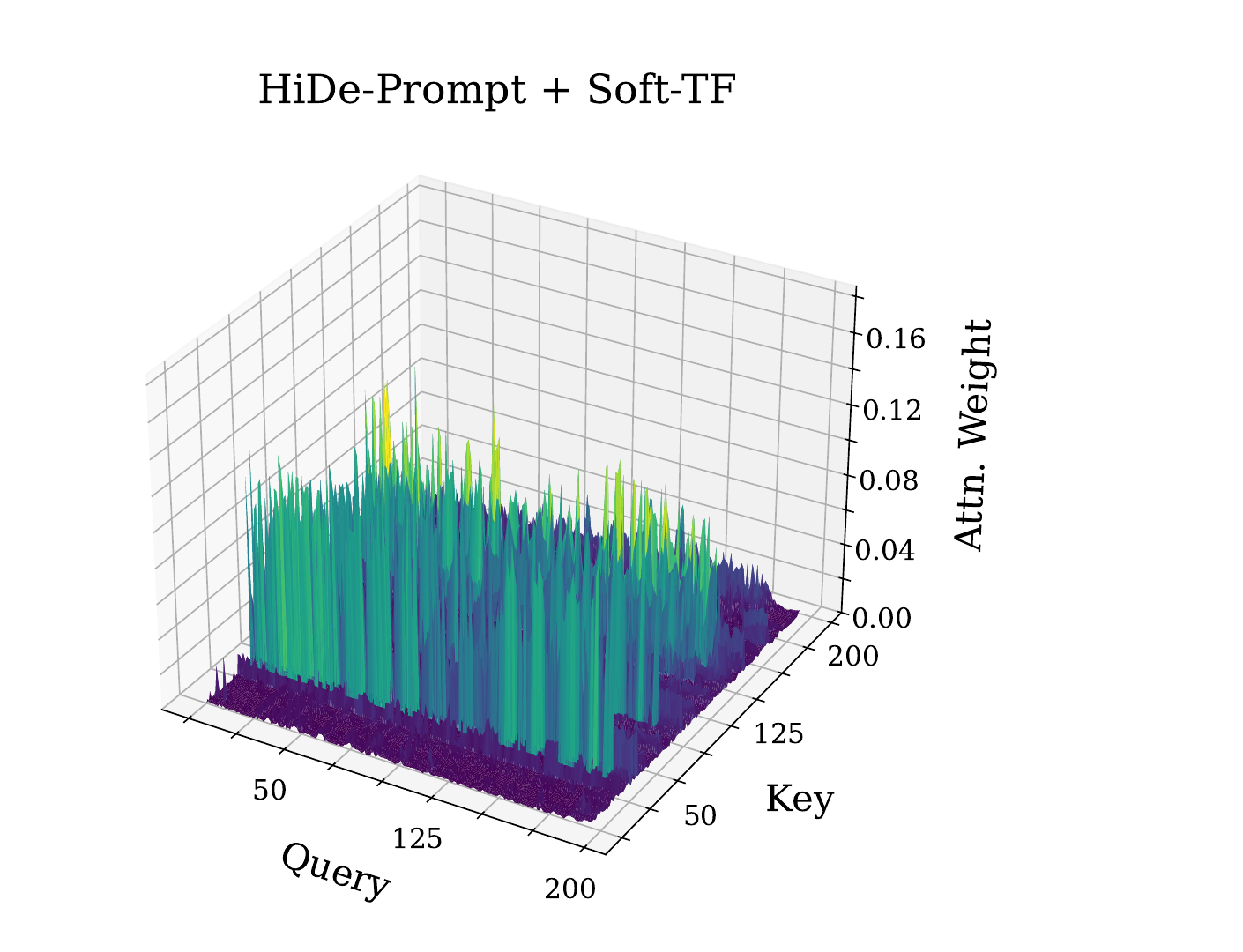} \\
    \small (a) HiDe-Prompt-L[12] & \small (b) HiDe-Prompt + Soft-TF-L[12] \\
    \end{tabular}
    }
    \caption{\small \textbf{Attention Layer-(L[12]) map comparison} of HiDe-Prompt and HiDe-Prompt + Soft-TF on a single 10-Split-CIFAR100 sample. The learned mask redistributes attention from a few dominant keys toward a broader task-relevant set, lowering the peak weight from $0.19$ to $0.12$ while sharing the same frozen backbone.}

    \label{fig:attn_l12_cifar10}
\end{figure}

\noindent
\textbf{Attention Layer.} We compare the 12th attention layer L[12] of HiDe-Prompt and HiDe-Prompt + Soft-TF on a single 10-Split-CIFAR100 sample, as shown in \Cref{fig:attn_l12_cifar10}. Both models share the same frozen backbone, so the only difference is the learned task mask. HiDe-Prompt concentrates attention on a few dominant key tokens with a sharp peak of $0.19$, whereas Soft-TF spreads it across a broader set of task-relevant tokens and lowers the peak to $0.12$. The mask therefore re-specializes attention toward the current task without displacing the pre-trained structure, in agreement with the near-identical weight distribution of \Cref{fig:soft_weight_cifar10} and the bounded-drift analysis of \Cref{sec:stability}. 

\noindent
\textbf{Additional Comparisons with Prompt-based Baselines.} As shown in \Cref{tab:app_prompt_perform}, on 10-Split-CIFAR100 the prompt learners L2P, DualPrompt, ESN, CODA-Prompt, and CPrompt obtain 77.14--86.61\% accuracy with 5.86--7.13 forgetting, whereas DualPrompt + Soft-TF reaches 92.35\% with 2.98 forgetting---$+5.74$\%p over the best baseline (DualPrompt) and roughly half its forgetting. Prompts condition each layer's input rather than its weights; Soft-TF instead modulates the attention projections directly, and this advantage persists even under random mask initialization (\Cref{tab:app_init_perform}).

\begin{table}[ht]
\begin{center}
\caption{\small \textbf{Comparison with prompt-based methods} on 10-Split-CIFAR100. DualPrompt + Soft-TF attains the highest accuracy and the lowest forgetting.}

\resizebox{0.32\textwidth}{!}{
\begin{tabular}{l r r}
\toprule
 &\multicolumn{2}{c}{\textbf{10-Split-CIFAR100}}   \\
\cmidrule(lr){2-3}
\textbf{Method} & ACC($\uparrow$) & Forget($\downarrow$)  \\ \midrule

L2P        & 83.38     &  5.88     \\
DualPrompt & 86.61     &  5.86     \\
ESN        & 86.42     &  6.08     \\
CODA-Prompt  & 85.73     &  7.13     \\ 
CPrompt & 77.14          & 5.97      \\
\rowcolor{gg} DualPrompt+\textbf{Soft-TF}   & \textbf{92.35} &  \textbf{2.98}    \\ 

\bottomrule
\end{tabular}}
\label{tab:app_prompt_perform}
\end{center}
\end{table}

\noindent
\textbf{Additional Results on CUB-200 \& 5-Datasets.}
As shown in \Cref{tab:fa_cub_5datasets}, adding Soft-TF to NoRGa raises final accuracy on both Split CUB-200 and 5-Datasets under Sup-21K and iBOT-21K pre-training. On Split CUB-200 it improves NoRGa from 90.90 to 97.90 under Sup-21K and from 80.69 to 90.82 under iBOT-21K; on 5-Datasets, from 94.16 to 95.68 and from 94.92 to 95.42. The gains hold across both pre-training regimes and are largest where the baseline is weakest ($+10.1$\%p on iBOT-21K CUB-200), consistent with weight-level adaptation reaching cases that input-level prompting alone cannot.

\begin{table}[t]
\centering
\caption{\small \textbf{Final average accuracy (FA)} on Split CUB-200 and 5-Datasets under Sup-21K and iBOT-21K pre-training. NoRGa + Soft-TF consistently improves over the prompt-based baselines.}
\label{tab:fa_cub_5datasets}
\resizebox{0.46\textwidth}{!}{
\begin{tabular}{l r r r r}
\toprule
\multirow{2}{*}{Method}
& \multicolumn{2}{c}{Split CUB-200}
& \multicolumn{2}{c}{5-Datasets} \\
\cmidrule(lr){2-3} \cmidrule(lr){4-5}
& Sup-21K & iBOT-21K & Sup-21K & iBOT-21K \\
\midrule
L2P          & 75.46 & 46.60 & 81.84 & 82.25 \\
DualPrompt  & 77.56 & 45.93 & 77.91 & 68.03 \\
S-Prompt    & 77.13 & 44.22 & 86.06 & 77.20 \\
CODA-Prompt & 74.34 & 47.79 & 64.18 & 51.65 \\
HiDe-Prompt & 86.56 & 78.23 & 93.83 & 94.88 \\
\midrule
NoRGa & {90.90} & {80.69} & {94.16} & {94.92} \\
\rowcolor{gg} NoRGa + \textbf{Soft-TF} & \textbf{97.90} & \textbf{90.82} & \textbf{95.68} & \textbf{95.42} \\
\bottomrule
\end{tabular}}
\end{table}

\section{Conclusion}\label{sec:conclusion}
Inspired by WLTH, we introduced Soft-TF, a continual learning framework that adapts a frozen pre-trained Transformer by learning task-specific soft subnetworks—real-valued multiplicative masks over the query, key, value, and output projections of selected self-attention layers. Initialized at one, the masks start optimization exactly at the pre-trained solution, and mask-space gradient descent is intrinsically biased toward modulating the backbone's dominant pathways; our analysis shows that both convergence speed and parameter drift are controlled by how close a task-optimal configuration lies to the pre-trained weights. This bounded drift underlies two defining behaviors: forgetting is structurally eliminated because masks are never overwritten, and class-incremental accuracy is decoupled from task-inference reliability because mis-selected masks stay near the generalist backbone---a selector correct on only 58\% of samples still yields near-87\% accuracy. Coupled with a dual-prompt mechanism and shared task-key retrieval, Soft-TF acts as a plug-in that consistently improves L2P, DualPrompt, HiDe-Prompt, and NoRGa, achieving state-of-the-art CIL performance across Split-CIFAR100, Split-ImageNet-R, CUB-200, and 5-Datasets while keeping inference cost identical to the unmodified backbone.

\noindent
\textbf{Limitations and Future Work.}
Soft-TF stores one mask set per task, so mask memory grows linearly with the number of tasks, adding $4D^2|\mathcal{L}_m|$ parameters per task; although masks can be kept in reduced precision and restricted to a few layers, very long task sequences motivate mask compression or consolidation. Our evaluation also focuses on ViT backbones and vision CIL benchmarks. Extending Soft-TF to broader architectures, modalities, and task-free settings, and pairing it with stronger task-inference selectors from which it benefits directly, are promising directions for future work.

\appendix
\section{Appendix}

\subsection{Convergence and Drift Analysis of Soft-TF}\label{app:convergence}
This appendix states and proves the result summarized in \Cref{thm:drift}. Throughout, $f$ denotes the task loss as a function of the effective weights $\bm{w} = \bm{\theta} \odot \bm{m}$ of \Cref{eq:effective}, and we work in \emph{mask coordinates}: define
\begin{equation}\label{eq:app_g}
g(\bm{m}) \;=\; f(\bm{\theta} \odot \bm{m}),
\end{equation}
where $\bm{\theta}$ is fixed. Since $\bm{m} \mapsto \bm{\theta} \odot \bm{m}$ is a linear map, $g$ is convex whenever $f$ is, and by the chain rule
\begin{equation}\label{eq:app_grad}
\nabla g(\bm{m}) \;=\; \bm{\theta} \odot \nabla_{\bm{w}} f(\bm{\theta} \odot \bm{m}).
\end{equation}
Soft-TF trains only the masks, i.e., it runs gradient descent on $g$ from the initialization $\bm{m}^{(1)} = \mathbf{1}$:
\begin{equation}\label{eq:app_update}
\bm{m}^{(k+1)} \;=\; \bm{m}^{(k)} - \eta\, \bm{v}_k,
\qquad
\bm{v}_k = \nabla g\big(\bm{m}^{(k)}\big).
\end{equation}

Our analysis rests on two explicitly stated assumptions. The first is a standard simplification—convexity does not hold globally for deep networks, and we use it only to obtain a transparent local analysis whose \emph{dependence on the initialization quality} is the object of interest. The second is our formalization of the Well-initialized Lottery Ticket Hypothesis (WLTH).

\begin{itemize}[leftmargin=*]
    \item \textbf{Assumption A1 (convexity and bounded gradients).} $g$ is convex, and $\|\bm{v}_k\| \le \rho$ for all iterates visited by \Cref{eq:app_update}.
    \item \textbf{Assumption A2 (well-initialized backbone; WLTH).} There exists a task-optimal mask $\bm{m}^{\ast} \in \operatorname*{argmin}_{\bm{m}} g(\bm{m})$ with
    $\|\mathbf{1} - \bm{m}^{\ast}\| \le B$ for a small constant $B$.
\end{itemize}
Assumption A2 states that some task-optimal solution is reachable by a \emph{mild rescaling} of the pre-trained weights. It is a property of the backbone--task pair, not of our algorithm; it is empirically supported by the observations that learned masks concentrate tightly around $\mathbf{1}$ in \Cref{fig:soft_weight_cifar10} and that stronger pre-training yields uniformly better continual learning performance in \Cref{tab:main_WLTH}. Note that no analogous guarantee is available to prompt-only adaptation: prompts modify each layer's \emph{input} rather than its weights, so when task adaptation requires modulation outside the span reachable by input conditioning, no mask-like ``identity initialization'' argument applies.

\begin{lemma}[telescoping bound]\label{lem:telescope}
For any update of the form \Cref{eq:app_update} with arbitrary vectors $\bm{v}_1, \dots, \bm{v}_K$ and any $\bm{m}^{\ast}$,
\begin{equation}\label{eq:app_lemma}
\sum_{k=1}^{K} \big\langle \bm{m}^{(k)} - \bm{m}^{\ast}, \bm{v}_k \big\rangle
\;\le\;
\frac{\|\bm{m}^{(1)} - \bm{m}^{\ast}\|^2}{2\eta}
+ \frac{\eta}{2}\sum_{k=1}^{K}\|\bm{v}_k\|^2 .
\end{equation}
\end{lemma}

\begin{proof}
By \Cref{eq:app_update}, $\bm{m}^{(k+1)} - \bm{m}^{\ast} = (\bm{m}^{(k)} - \bm{m}^{\ast}) - \eta \bm{v}_k$, hence
\begin{equation}\label{eq:app_expand}
\begin{aligned}
\|\bm{m}^{(k+1)} - \bm{m}^{\ast}\|^2
={}&
\|\bm{m}^{(k)} - \bm{m}^{\ast}\|^2 + \eta^2 \|\bm{v}_k\|^2 \\
&- 2\eta \big\langle \bm{m}^{(k)} - \bm{m}^{\ast}, \bm{v}_k \big\rangle .
\end{aligned}
\end{equation}
Rearranging,
\begin{equation}\label{eq:app_rearrange}
\begin{aligned}
\big\langle \bm{m}^{(k)} - \bm{m}^{\ast}, \bm{v}_k \big\rangle
={}&
\frac{\|\bm{m}^{(k)} - \bm{m}^{\ast}\|^2 - \|\bm{m}^{(k+1)} - \bm{m}^{\ast}\|^2}{2\eta} \\
&+ \frac{\eta}{2}\|\bm{v}_k\|^2 .
\end{aligned}
\end{equation}
Summing over $k = 1, \dots, K$, the first term telescopes to
$\big(\|\bm{m}^{(1)} - \bm{m}^{\ast}\|^2 - \|\bm{m}^{(K+1)} - \bm{m}^{\ast}\|^2\big)/(2\eta)$,
and dropping the non-negative subtracted term yields \Cref{eq:app_lemma}.
\end{proof}

\begin{theorem}[convergence of mask-only fine-tuning]\label{thm:app_convergence}
Under Assumptions A1--A2, run \Cref{eq:app_update} from $\bm{m}^{(1)} = \mathbf{1}$ for $K$ steps with $\eta = \frac{B}{\rho \sqrt{K}}$, and let $\bar{\bm{m}} = \frac{1}{K}\sum_{k=1}^{K} \bm{m}^{(k)}$. Then
\begin{equation}\label{eq:app_theorem}
g(\bar{\bm{m}}) - g(\bm{m}^{\ast})
\;\le\;
\frac{B \rho}{\sqrt{K}} .
\end{equation}
\end{theorem}

\begin{proof}
By Jensen's inequality and convexity of $g$,
\begin{equation}\label{eq:app_jensen}
g(\bar{\bm{m}}) - g(\bm{m}^{\ast})
\;\le\;
\frac{1}{K} \sum_{k=1}^{K} \big( g(\bm{m}^{(k)}) - g(\bm{m}^{\ast}) \big).
\end{equation}
Convexity further gives, for every $k$,
\begin{equation}\label{eq:app_convexity}
\begin{aligned}
g(\bm{m}^{(k)}) - g(\bm{m}^{\ast})
&\;\le\;
\big\langle \bm{m}^{(k)} - \bm{m}^{\ast}, \nabla g(\bm{m}^{(k)}) \big\rangle \\
&\;=\;
\big\langle \bm{m}^{(k)} - \bm{m}^{\ast}, \bm{v}_k \big\rangle .
\end{aligned}
\end{equation}
Combining \Cref{eq:app_jensen,eq:app_convexity} with \Cref{lem:telescope}, and using $\|\bm{m}^{(1)} - \bm{m}^{\ast}\| = \|\mathbf{1} - \bm{m}^{\ast}\| \le B$ and $\|\bm{v}_k\| \le \rho$,
\begin{equation}\label{eq:app_combine}
g(\bar{\bm{m}}) - g(\bm{m}^{\ast})
\;\le\;
\frac{B^2}{2\eta K} + \frac{\eta \rho^2}{2} .
\end{equation}
Substituting $\eta = \frac{B}{\rho\sqrt{K}}$ gives \Cref{eq:app_theorem}.
\end{proof}

\begin{corollary}[bounded parameter drift]\label{cor:app_drift}
Under the conditions of \Cref{thm:app_convergence}, every iterate satisfies
$\|\bm{m}^{(k)} - \bm{m}^{\ast}\| \le \sqrt{2}\,B$, and consequently the effective weights $\bm{w}^{(k)} = \bm{\theta} \odot \bm{m}^{(k)}$ remain within
\begin{equation}\label{eq:app_drift}
\|\bm{w}^{(k)} - \bm{\theta}\|
\;\le\;
\|\bm{\theta}\|_\infty \, \|\bm{m}^{(k)} - \mathbf{1}\|
\;\le\;
(1 + \sqrt{2})\, \|\bm{\theta}\|_\infty \, B
\end{equation}
of the pre-trained weights for all $k = 1, \dots, K{+}1$.
\end{corollary}

\begin{proof}
Since $\bm{m}^{\ast}$ minimizes $g$, \Cref{eq:app_convexity} implies $\langle \bm{m}^{(k)} - \bm{m}^{\ast}, \bm{v}_k \rangle \ge g(\bm{m}^{(k)}) - g(\bm{m}^{\ast}) \ge 0$. Then \Cref{eq:app_expand} gives
$\|\bm{m}^{(k+1)} - \bm{m}^{\ast}\|^2 \le \|\bm{m}^{(k)} - \bm{m}^{\ast}\|^2 + \eta^2 \rho^2$, and unrolling from $\bm{m}^{(1)}$,
\begin{equation}\label{eq:app_unroll}
\|\bm{m}^{(k)} - \bm{m}^{\ast}\|^2
\;\le\;
B^2 + K \eta^2 \rho^2
\;=\;
2B^2
\end{equation}
with $\eta = \frac{B}{\rho\sqrt{K}}$. By the triangle inequality,
$\|\bm{m}^{(k)} - \mathbf{1}\| \le \|\bm{m}^{(k)} - \bm{m}^{\ast}\| + \|\bm{m}^{\ast} - \mathbf{1}\| \le (1+\sqrt{2})B$, and
$\|\bm{w}^{(k)} - \bm{\theta}\| = \|\bm{\theta} \odot (\bm{m}^{(k)} - \mathbf{1})\| \le \|\bm{\theta}\|_\infty \|\bm{m}^{(k)} - \mathbf{1}\|$.
\end{proof}

The bounded drift of \Cref{cor:app_drift} has two continual-learning consequences, stated next.

\begin{proposition}[structural forget-freeness]\label{prop:forgetfree}
Fix a task $t$. In Soft-TF the backbone $\bm{\theta}$ is frozen and the task-specific parameters $(\bm{e}_t, \bm{m}_t)$ are not modified while later tasks $t{+}1, \dots, \mathcal{T}$ are learned. Hence, conditioned on a correct routing decision $\hat{t}=t$, the predictive function $f_{\bm{\theta}, \bm{g}, \bm{e}_t, \bm{m}_t}$ is invariant across all sessions $T \ge t$ up to the shared prompt $\bm{g}$, so the accuracy on task $t$ cannot decrease with $T$. Any residual per-task drop is therefore confined to $\bm{g}$ and the router.
\end{proposition}

\begin{proof}
Only $\bm{g}$, the router, and the current task's parameters are updated at session $T$; by construction $\bm{\theta}$, $\bm{e}_t$, and $\bm{m}_t$ are held fixed once task $t$ is learned. Thus $f_{\bm{\theta}, \bm{g}, \bm{e}_t, \bm{m}_t}(\bm{x})$ depends on $T$ only through $\bm{g}$, and with $\bm{g}$ fixed it is identical at sessions $t$ and $T$; the task-$t$ accuracy is unchanged, i.e., forgetting is zero.
\end{proof}

The remaining factor is the router. The next corollary shows that, thanks to bounded drift, a wrong routing decision is benign.

\begin{corollary}[proximity of task subnetworks and benign misrouting]\label{cor:app_proximity}
Let $\bm{w}_i = \bm{\theta}\odot\bm{m}_i$ and $\bm{w}_j = \bm{\theta}\odot\bm{m}_j$ be the converged effective weights of tasks $i$ and $j$, each satisfying \Cref{eq:app_drift} with a common bound $B$. Then
\begin{equation}\label{eq:app_pair}
\|\bm{w}_i - \bm{w}_j\| \;\le\; 2(1+\sqrt{2})\,\|\bm{\theta}\|_\infty\, B .
\end{equation}
If, in addition, the network output $\Phi(\bm{w};\bm{x})$ is $L$-Lipschitz in the masked weights on the region of interest, then for every input $\bm{x}$
\begin{align}
\|\Phi(\bm{w}_i;\bm{x}) - \Phi(\bm{w}_j;\bm{x})\| &\le 2L(1+\sqrt{2})\,\|\bm{\theta}\|_\infty\, B, \label{eq:app_output_pair}\\
\|\Phi(\bm{w}_j;\bm{x}) - \Phi(\bm{\theta};\bm{x})\| &\le L(1+\sqrt{2})\,\|\bm{\theta}\|_\infty\, B . \label{eq:app_output_gen}
\end{align}
\end{corollary}

\begin{proof}
\Cref{eq:app_pair} is the triangle inequality applied to \Cref{eq:app_drift}: $\|\bm{w}_i - \bm{w}_j\| \le \|\bm{w}_i - \bm{\theta}\| + \|\bm{\theta} - \bm{w}_j\|$. Composing $L$-Lipschitzness with \Cref{eq:app_pair} gives \Cref{eq:app_output_pair}, and with \Cref{eq:app_drift} gives \Cref{eq:app_output_gen}.
\end{proof}

\noindent
\textbf{Discussion.} \Cref{thm:app_convergence,cor:app_drift} make the informal claim of \Cref{thm:drift} precise: under Assumption A2, mask-only fine-tuning converges at rate $\mathcal{O}(B\rho/\sqrt{K})$ and stays in an $\mathcal{O}(\|\bm{\theta}\|_\infty B)$ neighborhood of the pre-trained solution. \Cref{prop:forgetfree} and \Cref{cor:app_proximity} then convert this into the two continual-learning properties we claim. Immutability of $\bm{\theta}$ and the stored masks makes forgetting structural rather than optimization-dependent. And since all task subnetworks share the same $\mathcal{O}(\|\bm{\theta}\|_\infty B)$ tube around $\bm{\theta}$, applying task $j$'s mask to a task-$i$ input perturbs the logits by only $\mathcal{O}(\|\bm{\theta}\|_\infty B)$ and leaves them near the generalist output $\Phi(\bm{\theta};\bm{x})$; when $B$ is small under WLTH, a mis-selected mask evaluates a near-generalist function that the shared classifier can still resolve, which is exactly why class-incremental accuracy decouples from task-inference accuracy in \Cref{sec:exp}.

\subsection{Experimental Details}\label{app:exp_details}
For fair comparisons with the baselines~\citep{wang2022learning,wang2022dualprompt}, we use ViT B/16~\citep{dosovitskiy2020image} pre-trained on ImageNet-21K as our image encoder, which is kept frozen during training. We train and test on a single Quadro RTX 8000-48GB GPU for baselines and our Soft-TransFormers with Adam optimizer with $\beta_1 = 0.9$ and $\beta_2=0.999$.

We follow standard prompt-based continual learning settings to validate our method's effectiveness. When comparing our approach with L2P and Soft-TransFormers on the 10/20-Split-CIFAR100 datasets, we train the network for 5 epochs with a batch size of 16 and set the prompt length to 5. For the 10-Split-ImageNet-R dataset, we use 50 epochs, a batch size of 16, and a prompt length of 30. In comparison with DualPrompt and Soft-TransFormers on the 10/20-Split-CIFAR100 dataset, we train the network for 20 epochs with a batch size of 24 and set the expert prompt length to 5. For the 10-Split-ImageNet-R dataset, we set the epochs to 50, the batch size to 24, and the expert prompt length to 20. Additionally, in all benchmark datasets, the general prompt length is set to 5, and the location inserted into the prompt is kept consistent.

\begin{figure*}[ht]
    \centering
    \small
    \setlength{\tabcolsep}{0pt}{%
    \begin{tabular}{cc}

    \includegraphics[width=0.95\columnwidth]{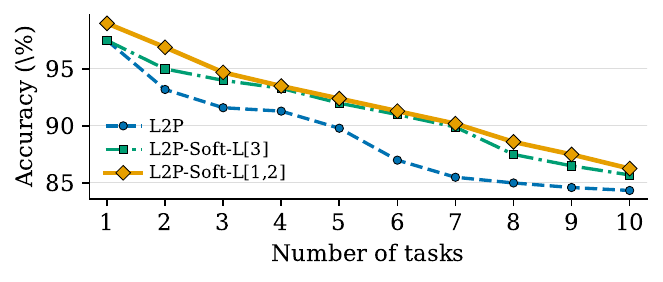} &

    \includegraphics[width=0.95\columnwidth]{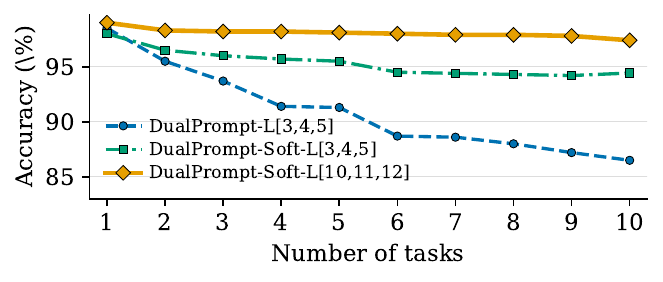} \\
    \small (a) L2P \textit{v.s.} L2P+Soft-TF & \small (b) DualPrompt \textit{v.s.} DualPrompt+Soft-TF \\
    
    \end{tabular}
    }
    \caption{\small \textbf{Layer-wise(L[$\ast$]) Performances of Soft-TF} on 10-Split-CIFAR100. Note that L[10,11,12] denotes Soft-TransFormer of 10, 11, 12 Layers.}

    \label{fig:layer_soft_cifar10}
    \vspace{-0.15in}
\end{figure*}

\noindent
\textbf{Layer-wise Inspections.} We analyze the layer-wise performance of Soft-TransFormers with respect to L2P and DualPrompt on the 10-Split-CIFAR100 dataset to identify the optimal configurations, as shown in \Cref{fig:layer_soft_cifar10}. Our observations reveal that the general prompt in DualPrompt influences Soft-TransFormers' performance differently in L2P and DualPrompt settings. In L2P, the best performance was achieved with Soft-TransFormers applied to the lower layers, shown as L2P-Soft-TF-L[1,2] in panel (a), whereas in DualPrompt the higher layers of DualPrompt-Soft-TF-L[10,11,12] in panel (b) yielded the best results. Notably, DualPrompt-Soft-TF-L[10,11,12] attains 0.21 forgetting, so masking the higher attention layers is the most effective configuration for the DualPrompt host.

\begin{figure}[ht]
    \centering
    \small
    \setlength{\tabcolsep}{-0.2pt}{%
    \begin{tabular}{cc}

    \includegraphics[width=0.5\columnwidth]{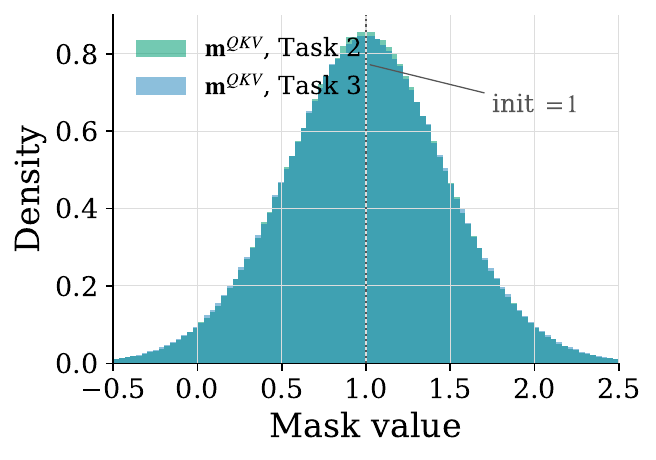} &

    \includegraphics[width=0.5\columnwidth]{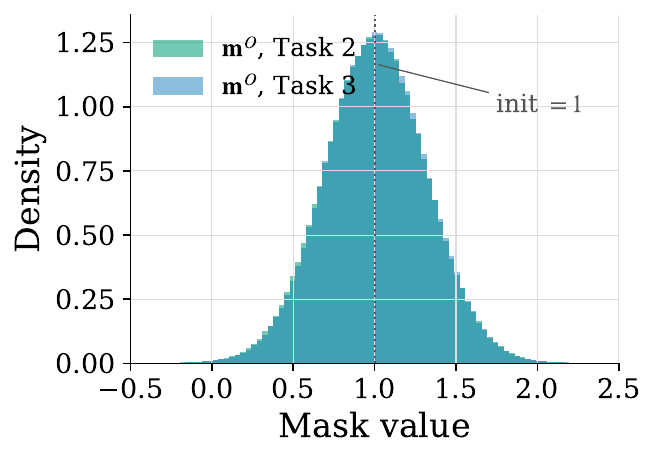} \\
    \small (a) Attn.QKV ($\bm{m}^{QKV}$) & \small (b) Attn.Proj ($\bm{m}^{O}$) \\

    \end{tabular}
    }
    \caption{\small \textbf{Layer-(L[12]) mask histogram density estimates} of HiDe-Prompt + Soft-TF on 10-Split-CIFAR100: (a) attention QKV mask $\bm{m}^{QKV}$ and (b) output-projection mask $\bm{m}^{O}$. Both stay centered at their initialization of $1$ and are nearly identical across Task 2 and Task 3.}

    \label{fig:app_layer_weight_cifar10}
    \vspace{-0.05in}
\end{figure}

\noindent \textbf{Analysis of Parameter Distributions.}
We examine the Layer-(L[12]) mask histograms of HiDe-Prompt + Soft-TF for the QKV attention (\textit{Attn.QKV}) and output projection (\textit{Attn.Proj}) modules, as illustrated in \Cref{fig:app_layer_weight_cifar10}. Both masks stay centered at their initialization of $1$ and are nearly identical across tasks. The \textit{Attn.QKV} masks exhibit a wider spread than the \textit{Attn.Proj} masks, indicating that the QKV projection absorbs most of the task-specific adaptation, so masking the QKV components is the most influential for Soft-TF.

\subsection{CLIP-Track Experiments: Isolating the Task-Inference Error}\label{app:clip}
Since Soft-TF is a plug-in over any frozen Transformer, we also evaluate it on a frozen CLIP ViT-B/16 image encoder~\cite{radford2021learning} on 10-Split-CIFAR100, as summarized in \Cref{tab:main_clip}. This CLIP track serves two purposes. First, it demonstrates backbone generality: coupling prompts with soft masks over all attention layers L[1-12] improves both CIL and TIL performance over CLIP-Prompt, reaching 96.83\% TIL accuracy with a near-zero forgetting of 0.44. Second, and more importantly, it isolates the task-inference error decomposition of \Cref{sub_sec:task_inference} in its purest form: for the same trained model, TIL results are identical under both task-inference mechanisms by construction, while the CIL accuracy jumps from 72.28\% with Prompt ID to 85.90\% with Gradient ID—a +13.6\%p gain obtained purely by improving task selection, with the representation untouched. This confirms that, once per-task soft subnetworks are stored, the residual CIL bottleneck lies in the selector rather than in the adapted representation, and that stronger inference mechanisms directly translate into end-to-end gains.

\begin{table}[hbt]
\begin{center}
\caption{\small \textbf{CLIP-track results} on 10-Split-CIFAR100 with a frozen CLIP ViT-B/16 encoder. For Soft-TF the CIL gap between Prompt ID and Gradient ID isolates the task-inference error.}
\resizebox{0.99\columnwidth}{!}{
\begin{tabular}{l c r r r r}
\toprule
 & & \multicolumn{2}{c}{\textbf{Class Incremental}} & \multicolumn{2}{c}{\textbf{Task Incremental}}  \\
\cmidrule(lr){3-4}\cmidrule(lr){5-6}
\multicolumn{1}{c}{\textbf{Method}} & \textbf{Task ID} & ACC($\uparrow$) & Forget($\downarrow$) & ACC($\uparrow$) & Forget($\downarrow$)  \\ \midrule
CLIP-Prompt                             & Prompt ID   & 70.27 & 12.95 & 93.36 & 3.07  \\
CLIP-Prompt-\textbf{Soft-TF}-L[3,4,5]   & Prompt ID   & 71.58 & 7.73  & 95.29 & 1.12  \\
CLIP-Prompt-\textbf{Soft-TF}-L[3,4,5]   & Gradient ID & 76.77 & 5.59  & 95.29 & 1.12  \\ \midrule
CLIP-Prompt-\textbf{Soft-TF}-L[1-12]    & Prompt ID   & 72.28 & 3.44  & 96.83 & \textbf{0.44}  \\
\rowcolor{gg} CLIP-Prompt-\textbf{Soft-TF}-L[1-12]    & Gradient ID & \textbf{85.90} & \textbf{3.07} & \textbf{96.83} & \textbf{0.44}  \\
\bottomrule
\end{tabular}}
\label{tab:main_clip}
\end{center}
\vspace{-0.1in}
\end{table}


\balance
\bibliographystyle{IEEEtranN}
\bibliography{reference}

%

\begin{IEEEbiography}[{\includegraphics[width=1in,height=1.25in,clip,keepaspectratio]{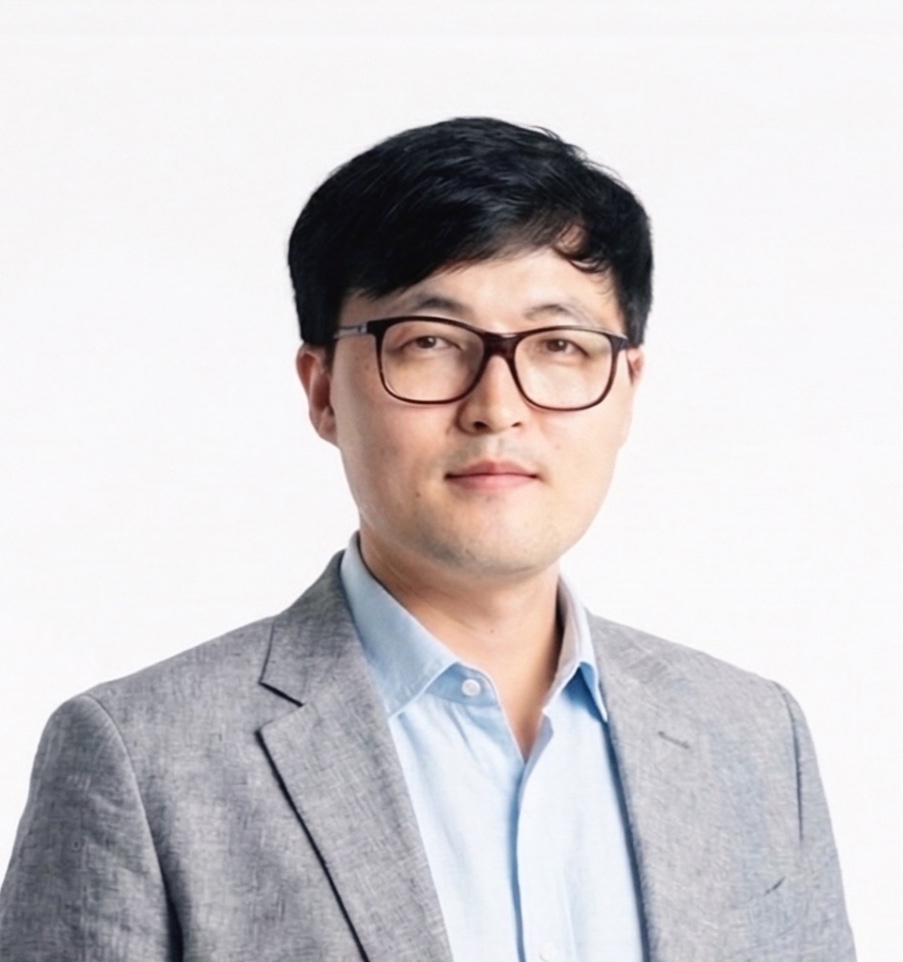}}]{Haeyong Kang}(Member, IEEE), (S'05) received the M.S. degree in Systems and Information Engineering from University of Tsukuba in 2007. From April 2007 to October 2010, he worked as an associate research engineer at LG Electronics. With working experiences at the Korea Institute of Science and Technology (KIST) and the University of Tokyo, He received a Ph.D. at the School of Electrical Engineering, the Korea Advanced Institute of Science and Technology (KAIST) with a dissertation on forget-free continual learning in 2023. He is pursuing research such as unbiased machine learning and continual learning as a postdoctoral researcher at KAIST.
\end{IEEEbiography}

\vspace{-13mm}

\begin{IEEEbiography}[{\includegraphics[width=1in,height=1.25in,clip,keepaspectratio]{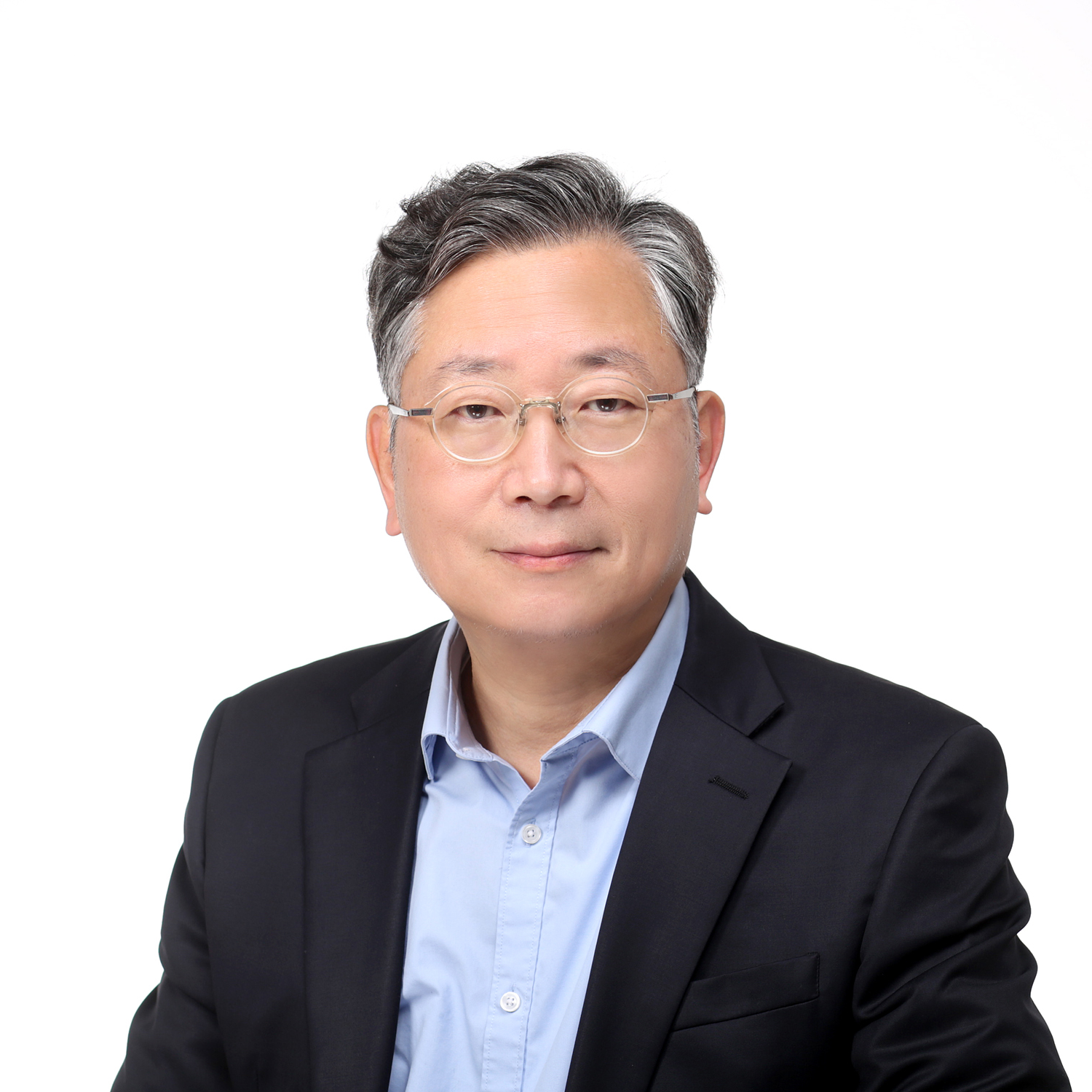}}]{Chang D. Yoo}
(Senior Member, IEEE) He received the B.S. degree in Engineering and Applied Science from the California Institute of Technology, the M.S. degree in Electrical Engineering from Cornell University, and the Ph.D. degree in Electrical Engineering from the Massachusetts Institute of Technology. From January 1997 to March 1999, he was Senior Researcher at Korea Telecom (KT). Since 1999, he has been on the faculty at the Korea Advanced Institute of Science and Technology (KAIST), where he is currently a Full Professor with tenure in the School of Electrical Engineering and an Adjunct Professor in the Department of Computer Science. He also served as Dean of the Office of Special Projects and Dean of the Office of International Relations.
\end{IEEEbiography}

\end{document}